\documentclass[sigconf]{acmart}

\usepackage{booktabs} % For formal tables
\usepackage{multirow}
\usepackage{epstopdf}

\setcopyright{rightsretained}
\begin{document}

\copyrightyear{2017}
\acmYear{2017}
\setcopyright{acmcopyright}
\acmConference{GECCO '17}{July 15-19, 2017}{Berlin, Germany}
\acmPrice{15.00}
\acmDOI{http://dx.doi.org/10.1145/3071178.3071328}
\acmISBN{978-1-4503-4920-8/17/07}

\title{Unsure When to Stop? Ask Your Semantic Neighbors}
%\title{Semantic Stopping Criteria and Their Application to Genetic Programming and Neural Networks}
%\titlenote{Produces the permission block, and
  %copyright information}
%\subtitle{Extended Abstract}
%\subtitlenote{The full version of the author's guide is available as
  %\texttt{acmart.pdf} document}

\author{Ivo Gon{\c{c}}alves}
%\authornote{This author did all the work. Literally all the work.}
\affiliation{
	\institution{NOVA IMS, Universidade Nova de Lisboa}%, Portugal}
	%\streetaddress{A}
	%\city{A} 
	%\state{A} 
	\postcode{1070-312 Lisbon, Portugal}
}
\email{igoncalves@novaims.unl.pt}

\author{Sara Silva}
\affiliation{
	\institution{BioISI - BioSystems \& Integrative Sciences Institute, Faculty of Sciences, University of Lisbon}%, Portugal}
	%\institution{BioISI, Faculty of Sciences, University of Lisbon}%, Portugal}
	%\streetaddress{A}
	%\city{A} 
	%\state{A} 
	\postcode{1749-016 Lisbon, Portugal}
}
\email{sara@fc.ul.pt}

\author{Carlos M. Fonseca}
\affiliation{
	\institution{CISUC, Department of Informatics Engineering, University of Coimbra}%, Portugal}
	%\streetaddress{A}
	%\city{A} 
	%\state{A} 
	\postcode{3030-290 Coimbra, Portugal}
}
\email{cmfonsec@dei.uc.pt}

\author{Mauro Castelli}
\affiliation{
	\institution{NOVA IMS, Universidade Nova de Lisboa}%, Portugal}
	%\streetaddress{A}
	%\city{A} 
	%\state{A} 
	\postcode{1070-312 Lisbon, Portugal}
}
\email{mcastelli@novaims.unl.pt}

\begin{abstract}

In iterative supervised learning algorithms it is common to reach a point in the search where no further induction seems to be possible with the available data. If the search is continued beyond this point, the risk of overfitting increases significantly. Following the recent developments in inductive semantic stochastic methods, this paper studies the feasibility of using information gathered from the semantic neighborhood to decide when to stop the search. Two semantic stopping criteria are proposed and experimentally assessed in Geometric Semantic Genetic Programming (GSGP) and in the Semantic Learning Machine (SLM) algorithm (the equivalent algorithm for neural networks). The experiments are performed on real-world high-dimensional regression datasets. The results show that the proposed semantic stopping criteria are able to detect stopping points that result in a competitive generalization for both GSGP and SLM. This approach also yields computationally efficient algorithms as it allows the evolution of neural networks in less than 3 seconds on average, and of GP trees in at most 10 seconds. The usage of the proposed semantic stopping criteria in conjunction with the computation of optimal mutation/learning steps also results in small trees and neural networks.

% This stochastic search can be performed over the space of Neural Networks or the space of tree-based models.

\end{abstract}

% ##### CCS classification #####
%%
%% The code below should be generated by the tool at
%% http://dl.acm.org/ccs.cfm
%\begin{CCSXML}
%<ccs2012>
%<concept>
%<concept_id>10010147.10010257.10010258.10010259</concept_id>
%<concept_desc>Computing methodologies~Supervised learning</concept_desc>
%<concept_significance>500</concept_significance>
%</concept>
%<concept>
%<concept_id>10010147.10010257.10010258.10010259.10010264</concept_id>
%<concept_desc>Computing methodologies~Supervised learning by regression</concept_desc>
%<concept_significance>500</concept_significance>
%</concept>
%<concept>
%<concept_id>10010147.10010257.10010293.10011809.10011813</concept_id>
%<concept_desc>Computing methodologies~Genetic programming</concept_desc>
%<concept_significance>500</concept_significance>
%</concept>
%</ccs2012>
%\end{CCSXML}
%
%\ccsdesc[500]{Computing methodologies~Supervised learning}
%\ccsdesc[500]{Computing methodologies~Supervised learning by regression}
%\ccsdesc[500]{Computing methodologies~Genetic programming}

% We no longer use \terms command

\keywords{Geometric Semantic Genetic Programming, Semantic Learning Machine, Generalization, Overfitting, Stopping Criteria}

\maketitle

% ####################################################################################################
%
% Section: Introduction
%
% ####################################################################################################
\section{Introduction}

%This also occurred in the early days of classical Machine Learning~\cite{Kushchu2002} [Domingos, 2012]. However, in the case of Machine Learning, the models used at the time were relatively simple. The underlying reasoning was that, since those models were simple, they were unlikely to overfit.
%... it is still understudied in comparison with other well-established non-evolutionary learning methods.
Supervised learning refers to the task of inducing a general pattern from a provided set of examples. A common issue in supervised learning is the possibility that the resulting models could be simply learning the provided set of examples, instead of learning the underlying pattern. A model that is incurring in such a behavior is commonly said to be overfitting. Genetic Programming (GP)~\cite{Koza1992} has been extensively applied in supervised learning tasks. Despite this, it was uncommon in the early years of GP to find approaches aimed at improving the generalization of the resulting models. Kushchu~\cite{Kushchu2002} mentioned that at that point, the issue of generalization in GP had not received the attention it deserved. In fact, it was even uncommon to measure the performance in a set of unseen data. Notably, in Koza~\cite{Koza1992} most of the problems presented did not use separate training and unseen data, so performance was never evaluated on unseen cases~\cite{Kushchu2002}. Eiben and Jelasity~\cite{Eiben2002GeccoEncoding} also mentioned that at that point, it was uncommon to report unseen data
results in the larger evolutionary computation area.

%has been recently increasing~\cite{Chen2016, Goncalves2015GSGP, Kommenda2014, Goncalves2013, Goncalves2012, Goncalves2011}. 
%Generalization in GP has been recently considered an important open issue~\cite{ONeill2010}, and the interest in studying generalization and overfitting has been recently increasing~\cite{GoncalvesPhdThesis, Chen2016, Goncalves2015GSGP, Kommenda2014, Goncalves2013, Goncalves2012, Goncalves2011}.
The interest in studying generalization and overfitting in GP has been recently increasing~\cite{GoncalvesPhdThesis, Chen2016, Goncalves2015GSGP, Kommenda2014, Goncalves2013, Goncalves2012, Goncalves2011}. Geometric Semantic Genetic Programming (GSGP)~\cite{Moraglio2012} has also contributed to this rising interest by defining a set of variation operators that have been shown to perform more effectively than the corresponding Standard GP operators~\cite{Moraglio2012, Goncalves2015GSGP}. These variation operators are known as geometric semantic operators. The reasoning behind these geometric semantic operators can be used to create equivalent operators for other representations or computational models. This was the case of feedforward neural networks, where Gon{\c{c}}alves et al.~\cite{Goncalves2015SLM} derived the original GSGP mutation operator to be applicable to neural networks. This operator was then incorporated in a neural network construction algorithm named Semantic Learning Machine (SLM). The SLM algorithm shares similar properties with GSGP. Within the context of these geometric semantic methods, this paper explores the feasibility of using information gathered with the geometric semantic operators to decide when to stop the search process. The selection of a suitable stopping point can potentially avoid the overfitting issue.

The paper is organized as follows. Section 2 contextualizes GSGP and the SLM. Section 3 presents the proposed semantic stopping criteria. Section 4 describes the experimental methodology. Section 5 presents and discusses the results, and section 6 concludes.

%number of publications dealing with generalization in GP has been increasing in the past few years~\cite{Goncalves2015GSGP,Goncalves2013V2,Goncalves2012, Goncalves2011, Castelli2010}. 
%Moraglio et al.~\cite{Moraglio2012}

% ####################################################################################################
%
% Section: Geometric Semantic Methods
%
% ####################################################################################################
\section{Geometric Semantic Methods}

\subsection{Geometric Semantic Genetic Programming}

% fitness landscape with a constant slope
%GSGP encompasses a set of new variation operators that share some specific characteristics.
The GSGP formulations are valid under two assumptions. The first assumption is that the task at hand is a supervised learning task, i.e., given a set of data with known targets, the goal is to build an individual (or model) that learns the underlying patterns in the data. The second assumption is that the error of an individual is computed as a distance to the known targets.
%This distance must be formally interpreted as a metric.
GSGP derives its name from the fact that its operators follow some particular geometric properties~\cite{MoraglioPhdThesis} over the semantic space. In this context, the semantics of an individual are defined as the outputs of that individual over a set of data instances. In GSGP, each individual is seen as a point in the semantic space.
%A geometric semantic crossover is an operator that always creates an offspring with semantics between the semantics of both parents. In other words, given any data instance, the output of the generated offspring must be between the outputs of the parents for that same data instance. A geometric semantic mutation is an operator that always creates an offspring with semantics that are within a given distance of the semantics of its parent. Depending on the version of the geometric semantic mutation, this distance may be possible to specify by a mutation step.
The most interesting property of the semantic space is that the associated fitness landscape is always unimodal for any supervised learning problem. 
%Formally, this fitness landscape is a metric cone.
This implies that there are no local optima, i.e., with the exception of the global optimum, every point in the search space has at least one neighbor with better fitness, and that neighbor is reachable through the application of the variation operators. As this type of landscape eliminates the local optima issue, it is potentially much more favorable in terms of search effectiveness and efficiency. Moraglio et al.~\cite{Moraglio2012} specified geometric semantic operators for three domains: boolean, arithmetic, and conditional rules. Since this paper is based on real-valued semantics, the geometric semantic operators presented here are the ones from the arithmetic domain.
% program (conditional rules)

\begin{definition} {\bf Geometric Semantic Crossover}:
Given two parent functions $T_1,T_2 : \mathbb{R}^n \to \mathbb{R}$, the geometric semantic crossover returns the real function $T_{XO}=(T_1 \cdot T_R) + ((1-T_R)\cdot T_2)$, where $T_R$ is a random real function whose output values range in the interval $[0,1]$.
\end{definition}

\begin{definition} {\bf Geometric Semantic Mutation}:
Given a parent function $T : \mathbb{R}^n \rightarrow \mathbb{R}$, the geometric semantic mutation with mutation step $ms$ returns the real function $T_M=T+ms\cdot(T_{R1} - T_{R2})$, where $T_{R1}$ and $T_{R2}$ are random real functions.
\label{def:gsgp-mutation}
\end{definition}

%The main drawback of the geometric semantic operators is that they always produce offspring bigger that their parents. Particularly for the crossover, every offspring is bigger than its parents combined. This leads to an exponential individual growth. For the mutation operator, the individual growth produced is linear.
An important consideration is that the geometric semantic mutation can reach any point in the semantic space, starting from any other point. On the other hand, in order for the geometric semantic crossover to produce an offspring which is better than both parents, the target semantics must be (at least partially) between the semantics of both parents. Theoretically, this means that the mutation operator should be more robust. This was confirmed in practice~\cite{Moraglio2012, Goncalves2015GSGP}. In the same works it was also empirically confirmed that a hill climbing strategy is indeed more efficient than a standard population-based strategy. Since there are no local optima, the search can be focused around the best individual without any conceivable disadvantage. Moraglio et al.~\cite{Moraglio2012} named this hill climbing strategy as Semantic Stochastic Hill Climber (SSHC). As a common hill climber, SSHC keeps only the best individual, and it uses only the geometric semantic mutation to advance the search process. At each generation a sample of offsprings is produced, and the new best is selected from the current best and the offsprings to survive to the next generation. SSHC is the GSGP variant used in this paper as it is more efficient than the standard population-based strategy.

An important question raised from the original GSGP work, was the issue of generalization. Moraglio et al.~\cite{Moraglio2012} did not measure the performance of the individuals in unseen data. Therefore, the generalization ability of the individuals produced by GSGP was unknown. Gon{\c{c}}alves et al.~\cite{Goncalves2015GSGP} showed that the resulting GSGP generalization is greatly dependent on a particular detail of the geometric semantic mutation. If the mutation operator generates $T_{R1}$ and $T_{R2}$ without any particular concern on the structure of the trees, the result is considerable overfitting. To differentiate, this mutation version was named unbounded mutation. On the other hand, if $T_{R1}$ and $T_{R2}$ are generated with a bounding function at the root node, the outcome is a competitive generalization. This mutation version was named bounded mutation. Applying a bounding function to $T_{R1}$ and $T_{R2}$ results in bounding the semantic variation also on unseen data. If a logistic function ($f(x) = \frac{1}{1 + e^{-x}}$) is used as a bounding function, then the output of each random subtree ranges in the interval $[0, 1]$ and, consequently, the output resulting from subtracting these random subtrees ranges in the interval $[-1, 1]$. After applying the mutation step, the semantic variation (or perturbation) added to the parent always ranges in the interval $[-ms, ms]$. By itself, this does not guarantee a competitive generalization. For a detailed discussion on this issue the reader is referred to~\cite{Goncalves2015GSGP}. Since the bounded mutation yields better generalizations, it is the one used in this paper.
%(or squashing)

%This continuous individual growth produced by the geometric semantic operators renders GSGP hard to use in practice, particularly in real-world multidimensional datasets. To counteract this growth, Moraglio et al.~\cite{Moraglio2012} proposed a simplification step after each operator application. This simplification approach was effective for the artificial problems tested, but it remains to be shown if it is effective in real-world multidimensional datasets.

\subsection{Semantic Learning Machine}

%In order to guarantee an equivalent comparison between the SLM and SSHC, the following considerations are adopted. 
The SLM~\cite{Goncalves2015SLM} is formulated under the same assumptions of GSGP. It also follows the same hill climbing strategy as SSHC. Similarly to the case of the GSGP mutation, the equivalent geometric semantic mutation for neural networks (NNs) is also a relatively simple combination of NNs. To simplify the description, the SLM mutation operator is referred to as GSM-NN, from Geometric Semantic Mutation - Neural Networks. This operator was originally specified for NNs with a single hidden layer~\cite{Goncalves2015SLM}, but it was subsequently extended to be applicable to any number of hidden layers~\cite{GoncalvesPhdThesis}. This paper only assesses the SLM in the single hidden layer variant. As such, the following GSM-NN description only applies to the single hidden layer case.

From an already existing NN (a parent NN), GSM-NN generates a random NN, and joins both in the same resulting NN. The first step of GSM-NN is to create the random NN. In this case, the random NN created is the simplest possible, i.e., it is a NN with a single hidden neuron. The weights from the input layer to the hidden neuron are randomly generated between -1.0 and 1.0 with uniform probability. The weight from the hidden neuron to the output neuron is the learning step ($ls$). This learning step is equivalent to the mutation step in the GSGP mutation. It influences the amount of semantic variation for each application of the operator. The second step of GSM-NN is to join the parent NN with the random NN. In this case of a single hidden layer NN, this join operation is rather simple since the parent NN and the random NN do not influence each other directly. Their semantics are simply combined at the output neuron to produce the final semantics resulting from the GSM-NN application. Because of this, GSM-NN can always perform an incremental evaluation at each mutation application. In other words, regardless of the size of a NN, the evaluation only needs to occur in the new part of the NN (the random NN). The contribution of the rest of the NN (the parent NN) is already computed, and therefore does not need to be reevaluated. This makes the SLM algorithm very efficient in practice.

In order to guarantee an equivalent comparison between the SLM and SSHC, all hidden neurons added in the application of GSM-NN have the hyperbolic tangent as their activation function. This guarantees that, for each application of the operator, the semantic variation always ranges in the interval $[-ls, ls]$, where $ls$ represents the learning step. This results from the fact that the outputs of the hyperbolic tangent range in the interval $[-1, 1]$. Similarly, in the SSHC variant considered here the semantic variation always ranges in the interval $[-ms, ms]$, where $ms$ represents the mutation step. It is important to remark that the GSM-NN operator excludes the need to use backpropagation to adjust the weights of the network. The GSM-NN operator allows the SLM algorithm to effectively explore the space of NNs. For further details the reader is referred to~\cite{Goncalves2015SLM} and~\cite{GoncalvesPhdThesis}.

\section{Semantic Stopping Criteria}

% Or: Empirical evidence shows that if the search is continued beyond this point
% Or: When the search is continued beyond this point, the risk of overfitting increases significantly.
The two semantic stopping criteria proposed here are intended to assess the feasibility of using the semantic neighborhood as a source of information to detect suitable stopping points in semantic supervised learning methods. A suitable stopping point is a point within the search where no further induction is possible with the available data. A common outcome of continuing the search beyond this point is overfitting the training data. In the best case scenario, the search enters generalization plateau in which the generalization achieved stabilizes. In either case, the best decision is to stop the search. Within this context, the term semantic neighborhood is used to define the set of models (neighbors) that are reachable from a given reference model when a given semantic mutation operator is applied. The reference model considered here is always the best model in terms of training data performance. A sampling of the semantic neighborhood is performed at each iteration/generation, and the decision of when to stop is based on the information of each semantic sample. Since the semantic methods considered in this paper follow a hill climbing strategy, a semantic sampling is already performed at each iteration/generation in order to advance the search. Therefore, the semantic stopping criteria studied here do not require any additional sampling. The underlying idea of these stopping criteria is to assess what the trend within the search is, and based on that decide a suitable stopping point that can avoid overfitting and/or reduce the computational time of the search method.

%... maybe mention that these criteria could be applied to Standard GP or other non-semantic methods

\subsection{Error Deviation Variation}

The Error Deviation Variation (EDV) criterion is based on the variation of the error deviation within the semantic neighborhood. In this context, the term error deviation is used to refer to the sample standard deviation of the absolute errors of a given model over the training instances. This criterion is only concerned with the models that improve over the current best model in a given iteration/generation. From these models, the criterion measures the percentage of models that reduce the error deviation in comparison with the error deviation of the current best model. In other words, from the neighbors that are better than the current best, the criterion measures the percentage of those that have a lower error deviation. This information allows to determine when the training error reduction starts to be conducted less uniformly across training instances. This may indicate that overfitting is starting to occur. The search is stopped when the criterion measure drops below a given stopping threshold (parameter). Since this criterion is based on the error deviation, it does not prevent the algorithm from finding models with a large semantic (output) deviation, as this may actually be desired given the target semantics. The criterion also does not prevent the next best model to have a larger error deviation than the previous best, as this flexibility could be important in the learning process. The search is only stopped if a considerable majority of the models are improving the training performance at the expense of larger error deviations.

\subsection{Training Improvement Effectiveness}

The Training Improvement Effectiveness (TIE) criterion is based on measuring the effectiveness of the semantic variation operator used to perform the sampling. In this context, the effectiveness of the operator is defined as the percentage of times that the operator is able to produce a neighbor that is superior to the current best model. During each iteration/generation, the effectiveness is measured with respect to the sample considered. As in the EDV criterion, the search is stopped when the effectiveness of the operator drops below a given stopping threshold. The reasoning underlying this criterion is that, if training error improvements are harder to find, then possibly these improvements are being forced at the expense of the resulting generalization.

% ####################################################################################################
%
% Section: Experimental Methodology
%
% ####################################################################################################
\section{Experimental Methodology}

The experimental methodology is based on Gon{\c{c}}alves et al.~\cite{Goncalves2015GSGP}, since this work has recently provided results for SSHC in two out of three datasets used in this paper. These datasets are the bioavailability (Bio), the plasma protein binding (PPB), and the median lethal dose (LD50). They have respectively: 359 instances and 241 features; 131 instances and 626 features; and 234 instances and 626 features. These real-world high-dimensional regression datasets describe relationships between pharmaceutical drugs and pharmacokinetics parameters.
%For a detailed description of these datasets the reader is referred to Archetti et al.~\cite{Archetti2007b}.

Table~\ref{tab:parameters} provides the experimental parameters. Furthermore, errors are computed as the Root Mean Squared Error (RMSE) between the outputs of a model and the targets of the dataset. The error on unseen data is referred to as generalization error. For each run a randomly selected data partition is performed. Each method uses the same data partition in equivalent runs. Notice that sample size refers to the number of models generated in the initialization and at each iteration/generation in SLM and SSHC. The term step is used as a simplification to refer to the learning step in SLM and the mutation step in SSHC. Two different step strategies are studied with the proposed stopping criteria: the Fixed Step (FS), and the Optimal Step (OS). As the name implies, FS variants (SLM-FS and SSHC-FS) use the same fixed step throughout the runs, while OS variants (SLM-OS and SSHC-OS) compute the optimal step for each application of the mutation operator by using the Moore-Penrose inverse. The computation of optimal steps under the SLM algorithm is explored for the first time in this paper. Gon{\c{c}}alves et al.~\cite{Goncalves2015SLM} only assessed the SLM algorithm under fixed steps. SLM-FS and SSHC-FS use a step of 1 for the Bio and PPB datasets (as in Gon{\c{c}}alves et al.~\cite{Goncalves2015GSGP}), and a step of 100 for the LD50 dataset. The higher step used in the LD50 dataset is explained by the higher order of magnitude of the errors. Claims of statistical significance are based on Mann-Whitney U tests, with Bonferroni correction, and considering a significance level of $\alpha = 0.05$. A non-parametric test is used because the data are not guaranteed to follow a normal distribution. All evolution plots presented in the next section are based on the median values over the 30 runs of some particular measure. The median is preferred over the average as it is more robust to outliers. Training error evolution plots are based on the training error of the best model at each generation. Generalization error evolution plots are based on the generalization error of the best model selected according to the training error.

% left out => State the usual Standard GP results for these datasets, and what a competitive RMSE generalization is under similar conditions for each dataset ... add references to these statements

\begin{table}
%\begin{table}[!htdp]
\caption{Parameters used in the experiments}
\label{tab:parameters} \centering
	%\begin{tabular}
	\resizebox{!}{18mm}{
	%\scalebox{0.75}{
	\begin{tabular}{@{\hspace{5mm}}l@{\hspace{5mm}}l@{\hspace{5mm}}}
		\hline \hline
		\textbf{Parameter} & \textbf{Value} \\ \hline \hline
		Data partition & Training 70\% - Unseen 30\% \\
		Runs & 30 \\
		Sample size & 100 \\
		SSHC initialization & Ramped Half-and-Half,\\ & maximum depth 6 \\
		SSHC function set & +, -, *, and / (protected) \\ % as in \cite{Poli2008}
		SSHC terminal set & Input variables, no constants \\
		%SSHC maximum tree depth & None \\
		%SSHC/SLM fixed step & Bio and PPB 1, LD50 100 \\
		\noalign{\smallskip} \hline \hline
	\end{tabular}}
\end{table}

\section{Experimental Study}
%\section{Results and Discussion}

% Or: Similar simplifications are used for SSHC: ...
To help the description the following simplifications are used: SLM-FS EDV describes SLM-FS using the EDV criterion; SLM-FS TIE describes SLM-FS using the TIE criterion; SLM-OS EDV describes SLM-OS using the EDV criterion; SSHC-FS EDV describes SSHC-FS using the EDV criterion; SSHC-FS TIE describes SSHC-FS using the TIE criterion; SSHC-OS EDV describes SSHC-OS using the EDV criterion. The TIE criterion is not applicable to the OS variants as the effectiveness of the mutation operator in this case can be empirically confirmed to be almost always 100\%. This means that no stopping occurs for OS variants (SLM-OS and SSHC-OS) under the TIE criterion for reasonable stopping thresholds.

\subsection{Semantic Learning Machine}

% ##### 1) criteria evolution without stopping just for SLM-FS #####

\begin{figure*}
%\begin{figure*}[!htdp]
	\begin{tabular}{ccc}
		\multicolumn{1}{c}{\textbf{Bio}}
		&
		\multicolumn{1}{c}{\textbf{PPB}}
		&
		\multicolumn{1}{c}{\textbf{LD50}}
		\\
		\\
		
		% HHH
		% 0.29
		%\includegraphics[width=0.5\textwidth, keepaspectratio]{figures/slm-fls-es-ssc-evolution-bio-sd}
		\includegraphics[height=50mm, width=0.29\textwidth, keepaspectratio]{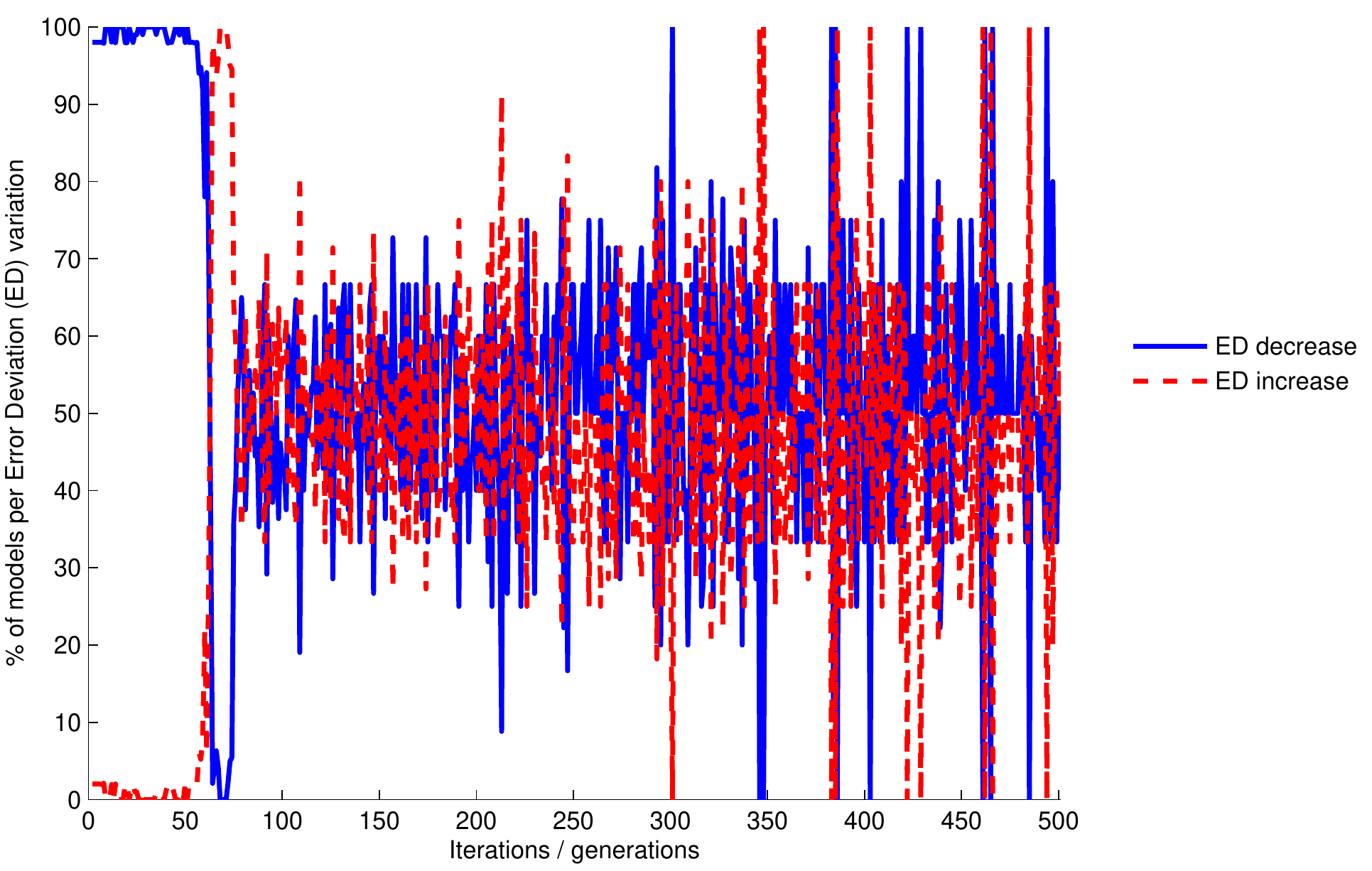}
		&
		\includegraphics[height=50mm, width=0.29\textwidth, keepaspectratio]{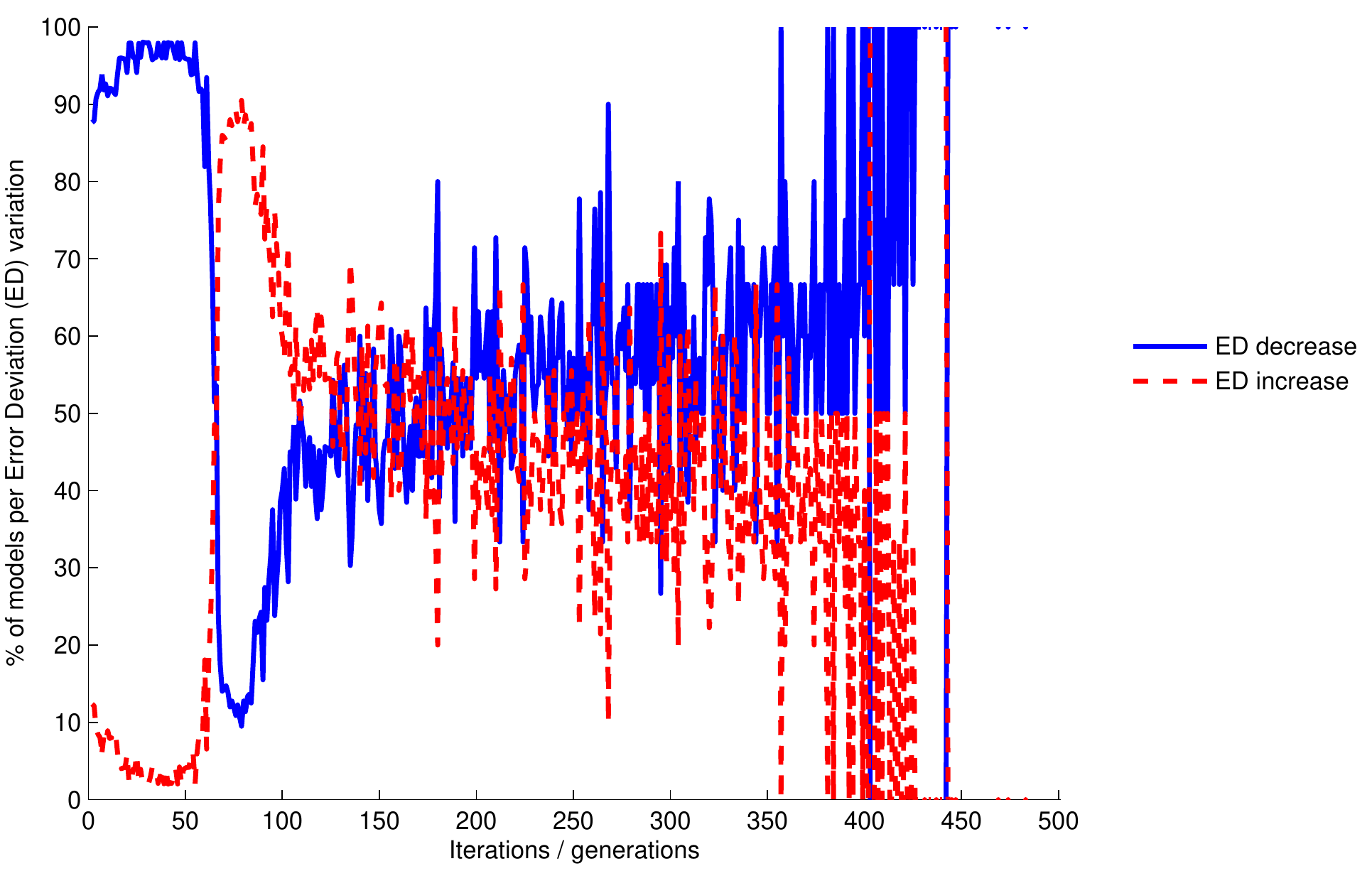}
		&
		\includegraphics[height=50mm, width=0.29\textwidth, keepaspectratio]{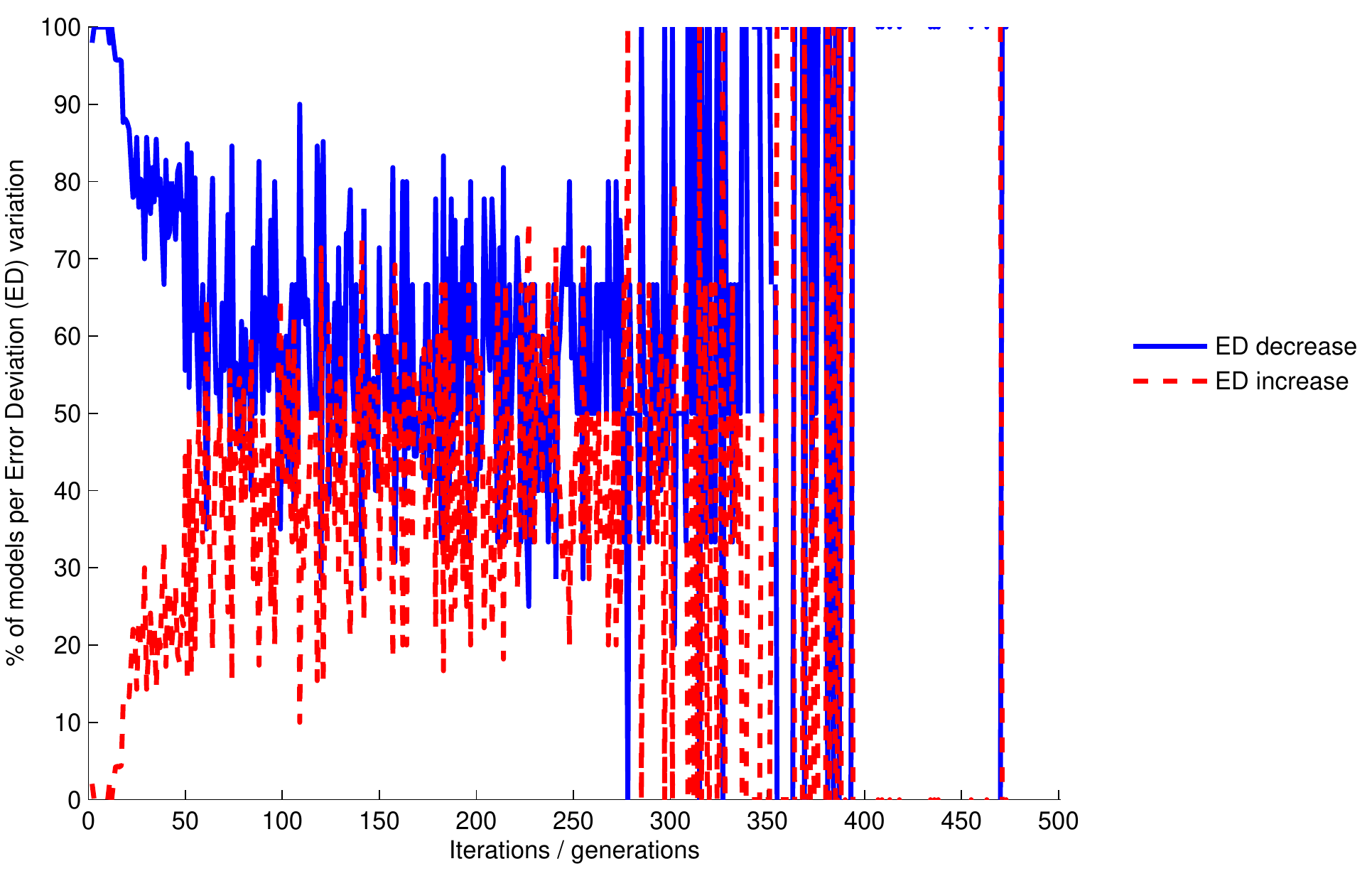}
		\\
		\includegraphics[height=50mm, width=0.29\textwidth, keepaspectratio]{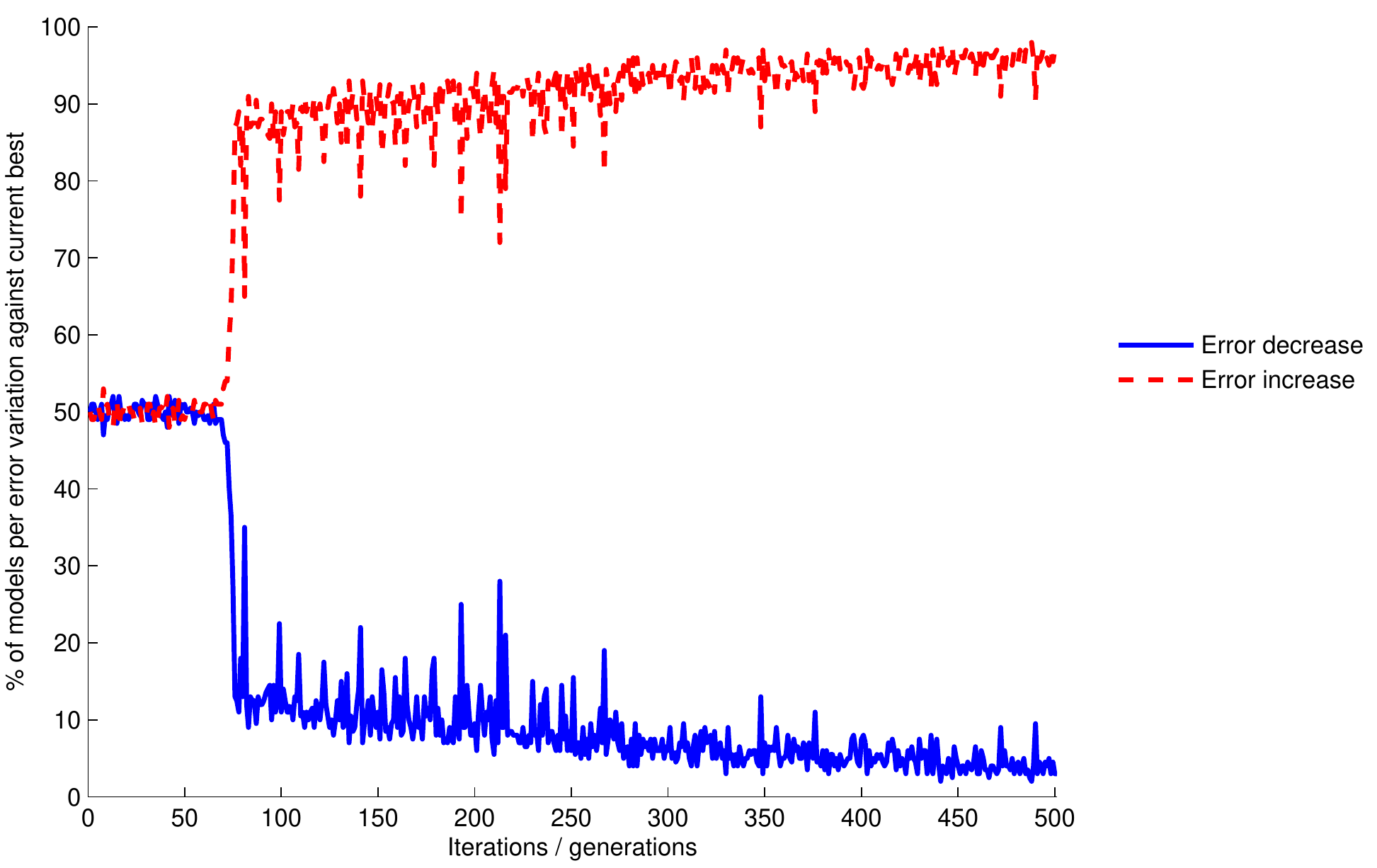}
		&
		\includegraphics[height=50mm, width=0.29\textwidth, keepaspectratio]{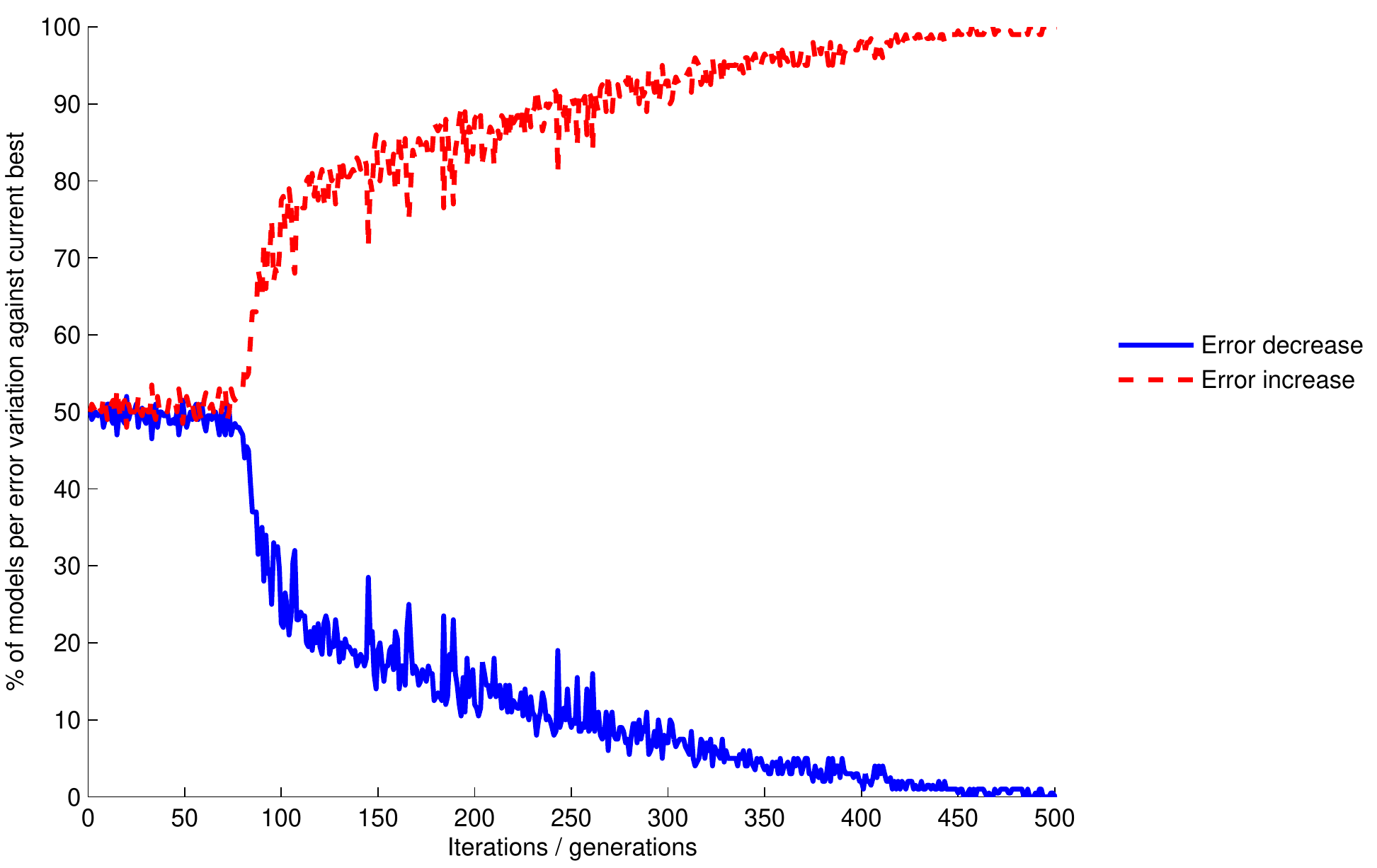}
		&
		\includegraphics[height=50mm, width=0.29\textwidth, keepaspectratio]{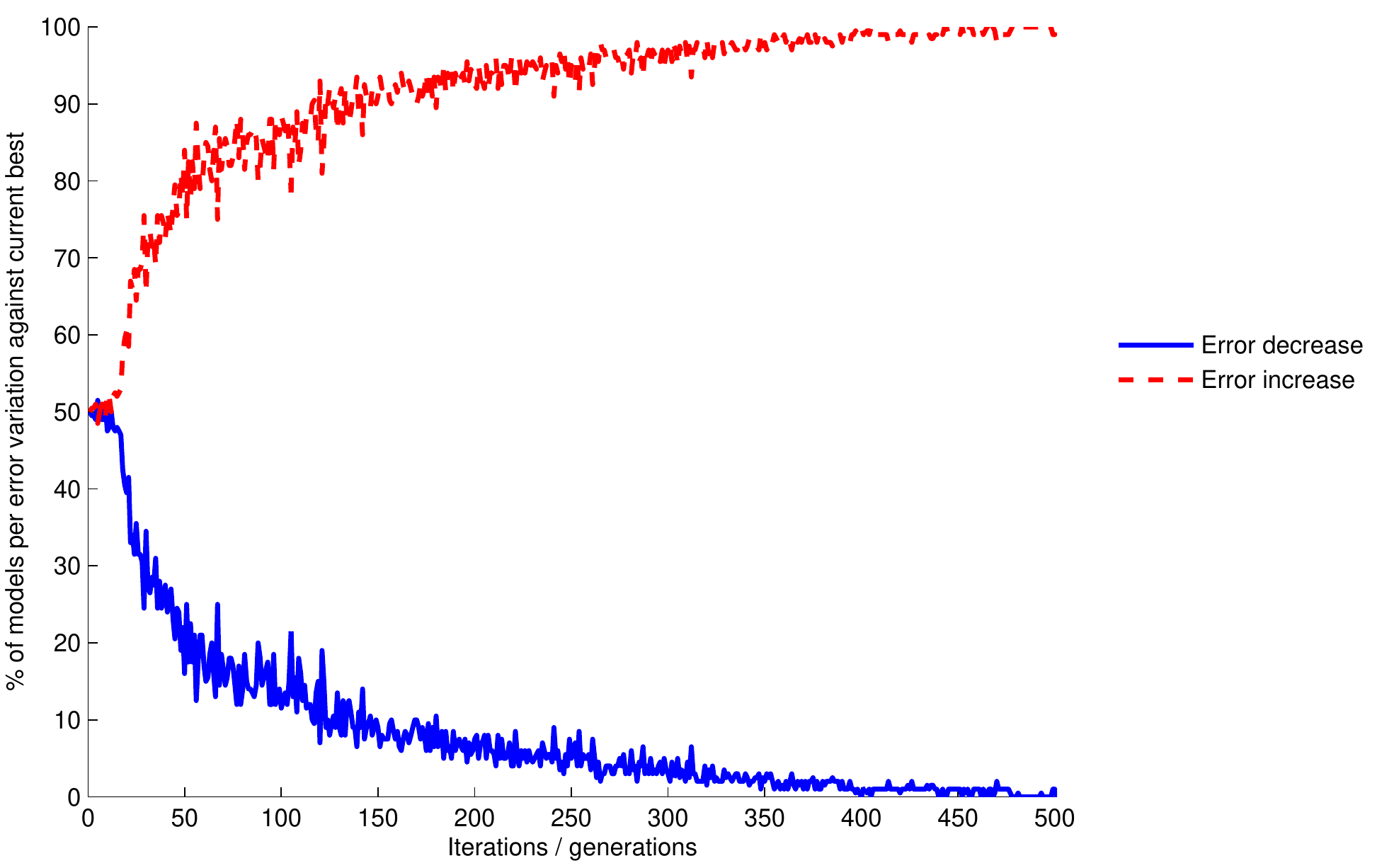}
	\end{tabular}
	\caption{Evolution of the measures related to the EDV (first row) and the TIE (second row) criteria for SLM-FS}
	\label{fig:slm-fls-es-ssc-evolution}
\end{figure*}

% Or: To exemplify the ...
As an initial assessment, figure~\ref{fig:slm-fls-es-ssc-evolution} presents the evolution of the measures used in the stopping criteria, and their complementary measures in SLM-FS. These plots show the medians of each value throughout the runs when the semantic stopping criteria are not applied. The first row presents the measures related with the EDV criterion, and the second row presents the measures related with the TIE criterion. The blue solid lines represent the measures that are directly used in the criteria to determine the stopping point. The red dashed lines are the respective complementary measures.

Starting with the EDV criterion (first row) in the Bio and PPB datasets, in the beginning of the runs the algorithm is very effective in generating models that are superior to the current best and, at the same time, reduce the error deviation (ED). This is shown by the line labeled as ED decrease. Until around iteration 50, the values of this measure are usually over 90\% in both datasets. Between iterations 50 and 100, a quick decrease of this measure occurs, as well as the consequent increase of the complementary measure (labeled as ED increase). The ED decrease measure drops below 20\%, while the ED increase measure increases to over 80\%. This indicates that the algorithm is mostly finding models that are superior to the current best at the expense of increasing the error deviation. It will be clear ahead that the area where this quick disruption occurs does indeed signal that no further induction seems to reliably possible.
%.. coincides with the generalization plateau, where the generalization error stabilizes while the training error keeps decreasing. This might be an interesting stopping point as no further induction seems to be possible.
%Continuing the search beyond this point might only increase the model size and the computational time.
In the LD50 dataset, the ED decrease measure also starts at very high values. However, a similar quick disruption is not as apparent from the median values presented. It will be clear ahead that a quick disruption also occurs in each run of the LD50 dataset. The fact that this disruption happens at considerably different iterations in different runs, makes the median values misleading.

In the TIE criterion (second row), a clear pattern occurs across all datasets. Initially, the effectiveness of the variation operator (labeled as error decrease) is around 50\%. In other words, generating a model which is superior to the current best is approximately as likely as generating a model which is inferior to the current best. This should be expected in the fixed step version of the mutation operator, as approximately 50\% of the models are generated in the direction of the target semantics, and the other 50\% are generated in the opposite direction. Similarly to what occurs in the EDV criterion, a disruption of this scenario happens between iterations 50 and 100 in the Bio and PPB datasets, and before iteration 30 in the LD50 dataset. Particularly noticeable in the Bio and PPB datasets, is the fact that this disruption in the TIE criterion occurs a few iterations later than in the EDV criterion.
%As in the EDV criterion, this disruption coincides with the generalization plateau.
After this disruption the effectiveness of the variation operator drops below 25\% across all datasets. The evolution of the measures used in the stopping criteria for the other variants tested (SLM-OS, SSHC-FS, and SSHC-OS) presents a similar behavior. These remaining results are not shown given the space restrictions.

% ##### 2) threshold robustness testing #####

% number of iterations conducted until the stopping occurred.
The next step is to assess the robustness of the stopping criteria with respect to the associated threshold. Intuitively, middle values (such as 25\%) should be more robust to sampling variations, while extreme stopping thresholds (5\% or 45\%) should lead to larger outcome variations. This paper considers 25\% to be the default stopping threshold for both criteria, while also experimentally assessing the outcomes for 15\% and 35\%. Table~\ref{tab:slm-table} presents the generalization errors, number of iterations, and training errors, resulting from the application of the stopping criteria to SLM-FS and SLM-OS with different stopping thresholds. Each performance outcome is presented with the median, average, and standard deviation (SD). These results show that the different stopping thresholds result in remarkably similar outcomes under each corresponding SLM variant. These results reflect very similar stopping points. This is a desirable outcome as no tuning of the stopping threshold seems to be required. Therefore, these stopping criteria are not simply shifting the burden of deciding when to stop, to a separate search over different threshold values. The remaining analysis considers the results with the default stopping threshold for both criteria (25\%).

%Iterations/generations
\begin{table*}
	\centering
	\caption{Median, average, and standard deviation (SD) results for generalization error, number of iterations, and training error, resulting from the application of the stopping criteria to SLM-FS and SLM-OS with different stopping thresholds}
	\label{tab:slm-table}
	\resizebox{!}{45mm}{
	%\scalebox{0.75}{
	\begin{tabular}{c|c|c|c|c|c|c|c|c|c|c|c}
		\multirow{2}{*}{\bf{Dataset}} & \multirow{2}{*}{\bf{SLM variant}} & \multirow{2}{*}{\bf{Threshold}} & \multicolumn{3}{c|}{\bf{Generalization error}} & \multicolumn{3}{c|}{\bf{Iterations}} & \multicolumn{3}{c}{\bf{Training error}} \\ \cline{4-12}
		& & & \bf{Median} & \bf{Average} & \bf{SD} & \bf{Median} & \bf{Average} & \bf{SD} & \bf{Median} & \bf{Average} & \bf{SD} \\ \hline
	
		\multirow{9}{*}{\bf{Bio}}

		& \multirow{3}{*}{\bf{FS EDV}}
		& 15\% & 31.0 & 30.9 & 1.0 & 61.5 & 60.8 & 2.3 & 30.0 & 30.1 & 0.9 \\ \cline{3-12}
		& & 25\% & 31.0 & 30.9 & 1.0 & 61.0 & 60.6 & 2.1 & 30.1 & 30.1 & 0.8 \\ \cline{3-12}
		& & 35\% & 31.0 & 30.9 & 1.0 & 61.0 & 60.6 & 2.1 & 30.1 & 30.1 & 0.8 \\ \cline{2-12}

		& \multirow{3}{*}{\bf{FS TIE}}
		& 15\% & 29.6 & 29.7 & 1.3 & 72.5 & 72.4 & 1.8 & 28.5 & 28.4 & 0.6 \\ \cline{3-12}
		& & 25\% & 29.6 & 29.7 & 1.3 & 72.0 & 72.3 & 1.7 & 28.5 & 28.4 & 0.6 \\ \cline{3-12}
		& & 35\% & 29.8 & 29.9 & 1.4 & 72.0 & 71.5 & 4.1 & 28.5 & 28.6 & 1.1 \\ \cline{2-12}

		& \multirow{3}{*}{\bf{OS EDV}}
		& 15\% & 30.1 & 30.2 & 1.6 & 3.0 & 3.4 & 1.2 & 29.2 & 29.2 & 0.6 \\ \cline{3-12}
		& & 25\% & 30.1 & 30.3 & 1.6 & 3.0 & 3.2 & 1.1 & 29.3 & 29.2 & 0.6 \\ \cline{3-12}
		& & 35\% & 30.1 & 30.3 & 1.6 & 3.0 & 3.1 & 1.1 & 29.3 & 29.3 & 0.7 \\ \hline

		\multirow{9}{*}{\bf{PPB}}

		& \multirow{3}{*}{\bf{FS EDV}}
		& 15\% & 31.1 & 31.1 & 1.7 & 66.5 & 66.0 & 2.9 & 28.2 & 28.1 & 1.2 \\ \cline{3-12}
		& & 25\% & 31.7 & 31.6 & 1.6 & 65.0 & 64.3 & 2.2 & 28.6 & 28.7 & 0.9 \\ \cline{3-12}
		& & 35\% & 31.7 & 31.9 & 1.8 & 65.0 & 63.4 & 2.8 & 28.8 & 29.1 & 1.2 \\ \cline{2-12}

		& \multirow{3}{*}{\bf{FS TIE}}
		& 15\% & 29.4 & 29.0 & 1.9 & 94.5 & 95.0 & 9.5 & 23.2 & 23.1 & 1.2 \\ \cline{3-12}
		& & 25\% & 29.1 & 28.9 & 1.8 & 85.0 & 86.2 & 4.5 & 23.9 & 23.9 & 1.0 \\ \cline{3-12}
		& & 35\% & 29.3 & 30.0 & 4.4 & 83.0 & 80.7 & 11.7 & 24.0 & 25.4 & 4.7 \\ \cline{2-12}

		& \multirow{3}{*}{\bf{OS EDV}}
		& 15\% & 31.8 & 32.6 & 5.1 & 7.5 & 9.6 & 6.9 & 25.0 & 24.6 & 2.5 \\ \cline{3-12}
		& & 25\% & 31.7 & 32.7 & 5.0 & 5.0 & 6.9 & 5.6 & 25.8 & 25.6 & 2.5 \\ \cline{3-12}
		& & 35\% & 32.0 & 32.7 & 4.9 & 4.5 & 6.7 & 5.7 & 25.8 & 25.8 & 2.7 \\ \hline

		\multirow{9}{*}{\bf{LD50}}

		& \multirow{3}{*}{\bf{FS EDV}}
		& 15\% & 2032.6 & 2058.0 & 187.8 & 27.5 & 30.5 & 9.3 & 1907.7 & 1907.8 & 109.8 \\ \cline{3-12}
		& & 25\% & 2028.6 & 2049.8 & 186.9 & 25.5 & 27.6 & 7.6 & 1940.0 & 1926.4 & 101.9 \\ \cline{3-12}
		& & 35\% & 2025.2 & 2049.3 & 187.3 & 25.0 & 26.7 & 7.1 & 1949.3 & 1932.0 & 99.4 \\ \cline{2-12}

		& \multirow{3}{*}{\bf{FS TIE}}
		& 15\% & 2051.3 & 2067.3 & 198.7 & 33.5 & 33.4 & 8.7 & 1908.7 & 1887.0 & 93.0 \\ \cline{3-12}
		& & 25\% & 2025.2 & 2036.8 & 189.8 & 22.0 & 22.2 & 2.4 & 1954.8 & 1962.0 & 79.9 \\ \cline{3-12}
		& & 35\% & 2016.5 & 2037.8 & 204.9 & 19.0 & 18.2 & 3.4 & 1995.2 & 2002.2 & 84.3 \\ \cline{2-12}

		& \multirow{3}{*}{\bf{OS EDV}}
		& 15\% & 2043.8 & 2067.1 & 201.4 & 5.0 & 6.0 & 3.8 & 1951.4 & 1940.2 & 90.3 \\ \cline{3-12}
		& & 25\% & 2013.8 & 2052.1 & 188.8 & 3.0 & 3.9 & 2.6 & 1980.4 & 1973.1 & 80.5 \\ \cline{3-12}
		& & 35\% & 1994.4 & 2014.7 & 203.3 & 1.0 & 2.2 & 2.2 & 2012.3 & 2006.5 & 82.3

	\end{tabular}}
\end{table*}

% ##### 3) all SLM variants with a 25\% stopping threshold #####

%(With respect to FS TIE vs FS EDV) The differences in the number of iterations conducted are also statistically significant ($p$-values: Bio $2.219\times10^{-11}$, and PPB $2.384\times10^{-11}$).

% Bio, generalization: FS TIE > FS EDV ($3.878\times10^{-4}$), FS TIE = OS EDV
% PPB, generalization: FS TIE > FS EDV ($9.899\times10^{-7}$), FS TIE > OS EDV ($3.274\times10^{-4}$)
% LD50, generalization: no differences

% LD50, iterations: FS EDV vs. FS TIE =>	s-	7.447540e-04
% LD50, iterations: FS EDV vs. OS EDV =>	s-	2.505473e-11
% LD50, iterations: FS TIE vs. OS EDV =>	s-	2.396206e-11

% Or: Overall, SLM-FS TIE yields the most robust generalizations.

%Figure~\ref{fig:slm-boxplots} presents the boxplots with the generalization errors (first row) achieved by applying the different stopping criteria to SLM-FS and SLM-OS. The same figure presents the number of iterations conducted (second row) until stopping occurred.
All variants avoid overfitting and result in competitive generalizations even though the OS and FS variants result in considerably different stopping points. As expected, the computation of optimal steps results in a considerably faster training error reduction. This then results in earlier stopping points across all datasets. SLM-OS EDV stops significantly faster than SLM-FS EDV ($p$-values: Bio $1.269\times10^{-11}$, PPB $2.367\times10^{-11}$, and LD50 $2.506\times10^{-11}$) and SLM-FS TIE ($p$-values: Bio $1.389\times10^{-11}$, PPB $2.611\times10^{-11}$, and LD50 $2.396\times10^{-11}$). In terms of generalization, the only case where a variant is significantly superior to the other two variants is in the PPB dataset, where SLM-FS TIE generalizes better than SLM-FS EDV ($p$-value $9.899\times10^{-7}$) and SLM-OS EDV ($p$-value $3.274\times10^{-4}$). Within the FS variants, SLM-FS TIE also generalizes better than SLM-FS EDV in the Bio dataset ($p$-value $3.878\times10^{-4}$). No other statistically significant differences are found in terms of generalization in the remaining comparisons. Overall, SLM-FS TIE is the most robust variant in terms of generalization. Interestingly, it almost always stops after the other variants. The only case where this does not happen is in the LD50 dataset, where it stops at a similar point as SLM-FS EDV.

\subsection{Semantic Stochastic Hill Climber}
%\subsection{Geometric Semantic Genetic Programming}

% ##### 1) threshold robustness testing #####

Similarly to table~\ref{tab:slm-table}, table~\ref{tab:sshc-table} presents the generalization errors, number of generations, and training errors, resulting from the application of the stopping criteria to SSHC-FS and SSHC-OS with the different stopping thresholds. The results are relatively consistent in avoiding overfitting, although they present a slightly bigger variation in comparison with the SLM results. Some negative outliers exist in terms of generalization, particularly in SSHC-OS EDV. For instance, in the LD50 dataset, the average and standard deviation are considerably influenced by a single run where the generalization error is over 70000. Without this outlier, the average generalization error for the 25\% threshold would be 2135.7, and the standard deviation would be 357.5. This behavior might be related with the semantic distributions generated by the random tree initializations. As empirically shown by Gon{\c{c}}alves et al.~\cite{Goncalves2015SLM}, even for otherwise equivalent algorithms, the random initializations used within the geometric semantic mutation operator can significantly influence the outcomes. This was also discussed by Moraglio et al.~\cite{Moraglio2012}. A further investigation on the influence of different random tree initializations within GSGP/SSHC might clarify this issue. The remaining analysis considers the results with the default stopping threshold for both criteria (25\%).

%Iterations/generations
\begin{table*}
	\centering
	\caption{Median, average, and standard deviation (SD) results for generalization error, number of generations, and training error, resulting from the application of the stopping criteria to SSHC-FS and SSHC-OS with different stopping thresholds}
	\label{tab:sshc-table}
	\resizebox{!}{45mm}{
	%\scalebox{0.75}{
	\begin{tabular}{c|c|c|c|c|c|c|c|c|c|c|c}
		\multirow{2}{*}{\bf{Dataset}} & \multirow{2}{*}{\bf{SSHC variant}} & \multirow{2}{*}{\bf{Threshold}} & \multicolumn{3}{c|}{\bf{Generalization error}} & \multicolumn{3}{c|}{\bf{Generations}} & \multicolumn{3}{c}{\bf{Training error}} \\ \cline{4-12}
		& & & \bf{Median} & \bf{Average} & \bf{SD} & \bf{Median} & \bf{Average} & \bf{SD} & \bf{Median} & \bf{Average} & \bf{SD} \\ \hline
		
		\multirow{9}{*}{\bf{Bio}}

		& \multirow{3}{*}{\bf{FS EDV}}
		& 15\% & 30.7 & 34.6 & 19.9 & 1029.5 & 1137.6 & 441.2 & 18.2 & 18.8 & 2.2 \\ \cline{3-12}
		%& 15\% & 30.7 & 34.6 & 19.9 & 1000.0 & 895.3 & 163.5 & 18.2 & 18.8 & 2.2 \\ \cline{3-12}
		& & 25\% & 33.7 & 37.5 & 19.6 & 378.5 & 344.5 & 281.4 & 25.0 & 29.0 & 7.7 \\ \cline{3-12}
		& & 35\% & 38.0 & 41.3 & 18.0 & 37.0 & 55.5 & 67.8 & 38.2 & 37.5 & 3.9 \\ \cline{2-12}

		& \multirow{3}{*}{\bf{FS TIE}}
		& 15\% & 30.7 & 34.6 & 19.9 & 1364.0 & 1358.7 & 133.7 & 17.6 & 17.6 & 0.8 \\ \cline{3-12}
		%& 15\% & 30.7 & 34.6 & 19.9 & 1000.0 & 999.4 & 3.3 & 17.6 & 17.6 & 0.8 \\ \cline{3-12}
		& & 25\% & 30.5 & 34.4 & 19.3 & 519.0 & 512.8 & 98.4 & 23.2 & 23.6 & 1.6 \\ \cline{3-12}
		& & 35\% & 35.4 & 40.1 & 19.1 & 99.5 & 97.7 & 54.2 & 33.6 & 35.6 & 7.2 \\ \cline{2-12}

		& \multirow{3}{*}{\bf{OS EDV}}
		%& 15\% & 44.7 & $7.9\times10^{7}$ & $4.1\times10^{8}$ & 369.0 & 313.5 & 197.2 & 9.8 & 13.6 & 6.9 \\ \cline{3-12}
		& 15\% & 44.7 & 7.9e+07 & 4.1e+08 & 369.0 & 313.5 & 197.2 & 9.8 & 13.6 & 6.9 \\ \cline{3-12}
		%& 15\% & 44.7 & 78792286.5 & 409242131.4 & 369.0 & 313.5 & 197.2 & 9.8 & 13.6 & 6.9 \\ \cline{3-12}
		& & 25\% & 33.6 & 231.6 & 1031.8 & 15.5 & 24.4 & 33.8 & 27.7 & 28.2 & 5.9 \\ \cline{3-12}
		& & 35\% & 34.9 & 38.5 & 20.2 & 2.0 & 3.2 & 2.2 & 33.7 & 33.5 & 3.5 \\ \hline

		\multirow{9}{*}{\bf{PPB}}

		& \multirow{3}{*}{\bf{FS EDV}}
		& 15\% & 32.0 & 32.5 & 4.0 & 218.5 & 228.5 & 165.1 & 20.9 & 21.4 & 7.9 \\ \cline{3-12}
		& & 25\% & 34.7 & 35.4 & 4.4 & 37.5 & 65.1 & 67.8 & 32.1 & 31.6 & 4.3 \\ \cline{3-12}
		& & 35\% & 36.8 & 37.2 & 5.3 & 18.5 & 29.7 & 29.8 & 33.9 & 35.0 & 3.7 \\ \cline{2-12}

		& \multirow{3}{*}{\bf{FS TIE}}
		& 15\% & 31.5 & 31.6 & 3.4 & 522.0 & 505.0 & 71.9 & 8.3 & 8.8 & 2.4 \\ \cline{3-12}
		& & 25\% & 31.6 & 32.1 & 3.7 & 240.5 & 244.6 & 73.4 & 18.4 & 19.5 & 3.0 \\ \cline{3-12}
		& & 35\% & 35.1 & 36.4 & 5.8 & 41.0 & 45.0 & 26.1 & 33.3 & 33.1 & 4.4 \\ \cline{2-12}

		& \multirow{3}{*}{\bf{OS EDV}}
		& 15\% & 35.1 & 62.0 & 147.6 & 5.0 & 9.0 & 11.3 & 28.4 & 27.0 & 5.9 \\ \cline{3-12}
		& & 25\% & 35.4 & 36.0 & 4.3 & 2.0 & 2.2 & 1.7 & 31.3 & 32.8 & 5.2 \\ \cline{3-12}
		& & 35\% & 36.7 & 37.4 & 5.4 & 1.0 & 1.1 & 1.2 & 33.6 & 35.7 & 6.3 \\ \hline

		\multirow{9}{*}{\bf{LD50}}

		& \multirow{3}{*}{\bf{FS EDV}}
		& 15\% & 2070.4 & 2106.5 & 165.4 & 143.0 & 148.0 & 66.2 & 1585.9 & 1570.9 & 204.4 \\ \cline{3-12}
		& & 25\% & 2041.5 & 2064.3 & 169.8 & 68.5 & 67.5 & 28.1 & 1814.6 & 1829.7 & 179.3 \\ \cline{3-12}
		& & 35\% & 2053.1 & 2063.1 & 193.9 & 33.0 & 34.8 & 15.8 & 1933.7 & 1971.1 & 158.6 \\ \cline{2-12}

		& \multirow{3}{*}{\bf{FS TIE}}
		& 15\% & 2183.3 & 2189.0 & 169.6 & 261.5 & 270.9 & 66.3 & 1268.5 & 1249.6 & 136.2 \\ \cline{3-12}
		& & 25\% & 2017.9 & 2059.9 & 168.1 & 81.0 & 82.5 & 31.5 & 1761.1 & 1768.5 & 87.8 \\ \cline{3-12}
		& & 35\% & 2044.3 & 2057.8 & 204.3 & 23.0 & 20.8 & 7.0 & 2025.7 & 2043.8 & 138.2 \\ \cline{2-12}

		& \multirow{3}{*}{\bf{OS EDV}}
		& 15\% & 2120.8 & 5463.5 & 13499.0 & 5.5 & 9.4 & 10.4 & 1875.1 & 1836.4 & 194.7 \\ \cline{3-12}
		& & 25\% & 2065.3 & 4402.9 & 12423.1 & 2.0 & 3.7 & 3.7 & 1972.9 & 1957.5 & 173.1 \\ \cline{3-12}
		& & 35\% & 2065.3 & 4332.3 & 12433.3 & 2.0 & 2.2 & 1.4 & 1989.7 & 2004.9 & 158.7

	\end{tabular}}
\end{table*}

As in the case of SLM, the computation of optimal steps in SSHC also results in earlier stopping points across all datasets. SSHC-OS EDV stops significantly faster than SSHC-FS EDV ($p$-values: Bio $4.272\times10^{-8}$, PPB $3.442\times10^{-10}$, and LD50 $4.486\times10^{-10}$) and SSHC-FS TIE ($p$-values: Bio $2.836\times10^{-11}$, PPB $2.638\times10^{-11}$, and LD50 $2.235\times10^{-11}$). In the Bio and PPB datasets, SSHC-FS TIE achieves significantly superior generalizations than SSHC-FS EDV ($p$-values: Bio $5.445\times10^{-3}$, and PPB $3.585\times10^{-3}$) and SSHC-OS EDV ($p$-values: Bio $7.476\times10^{-6}$, and PPB $6.728\times10^{-4}$). No statistically significant differences are found in terms of generalization in the LD50 dataset. Across all datasets, the best SSHC variant always achieves a competitive generalization. As in the SLM case, the TIE criterion applied in conjunction with a fixed step yields the most robust generalizations. These generalizations also result from later stops as in the SLM case.

%\begin{figure*}
%%\begin{figure*}[!htdp]
%%\begin{figure*}[d]
	%\begin{tabular}{ccc}
		%\multicolumn{1}{c}{\textbf{Bio}}
		%&
		%\multicolumn{1}{c}{\textbf{PPB}}
		%&
		%\multicolumn{1}{c}{\textbf{LD50}}
		%\\
		%\\
		%\includegraphics[height=50mm, width=0.32\textwidth, keepaspectratio]{figures/sshc-bm-and-abm-bio-generalization-boxplot}
		%&
		%\includegraphics[height=50mm, width=0.32\textwidth, keepaspectratio]{figures/sshc-bm-and-abm-ppb-generalization-boxplot}
		%&
		%\includegraphics[height=50mm, width=0.32\textwidth, keepaspectratio]{figures/sshc-bm-and-abm-ld50-generalization-boxplot}
		%\\
		%\includegraphics[height=50mm, width=0.32\textwidth, keepaspectratio]{figures/sshc-bm-and-abm-bio-generations-boxplot}
		%&
		%\includegraphics[height=50mm, width=0.32\textwidth, keepaspectratio]{figures/sshc-bm-and-abm-ppb-generations-boxplot}
		%&
		%\includegraphics[height=50mm, width=0.32\textwidth, keepaspectratio]{figures/sshc-bm-and-abm-ld50-generations-boxplot}
	%\end{tabular}
	%\caption{Boxplots with the generalization errors (first row), and the number of generations (second row) resulting from the application of the stopping criteria to SSHC-FS and SSHC-OS with a 25\% stopping threshold}
	%\label{fig:sshc-boxplots}
%\end{figure*}

\subsection{Additional Considerations}

With respect to the direct comparisons between equivalent SLM and SSHC variants when using the same stopping criterion, SLM-FS EDV generalizes better than SSHC-FS EDV in the Bio ($p$-value $2.211\times10^{-3}$) and PPB ($p$-value $5.715\times10^{-4}$) datasets. SLM-FS TIE generalizes better than SSHC-FS TIE in the PPB dataset ($p$-value $2.759\times10^{-4}$), and SLM-OS EDV generalizes better than SSHC-OS EDV in the Bio ($p$-value $1.256\times10^{-8}$) and PPB ($p$-value $1.723\times10^{-3}$) datasets. No other statistically significant differences are found in terms of generalization in the remaining equivalent comparisons. In general, these results show that the SLM variants are more robust than the equivalent SSHC variants. As previously mentioned, this might be related with the possibly less smooth semantic distributions generated by the GP random tree initializations.

\begin{figure*}
%\begin{figure*}[!htdp]
	\begin{tabular}{ccc}
		\multicolumn{1}{c}{\textbf{Bio}}
		&
		\multicolumn{1}{c}{\textbf{PPB}}
		&
		\multicolumn{1}{c}{\textbf{LD50}}
		\\
		\\
		%\includegraphics[height=100mm, width=0.32\textwidth, keepaspectratio]{figures/comparison-plot-bio-generalization}
		%&
		%\includegraphics[height=100mm, width=0.32\textwidth, keepaspectratio]{figures/comparison-plot-ppb-generalization}
		%&
		%\includegraphics[height=100mm, width=0.32\textwidth, keepaspectratio]{figures/comparison-plot-ld50-generalization}

		% HHH
		% 0.32
		\includegraphics[width=0.31\textwidth, keepaspectratio]{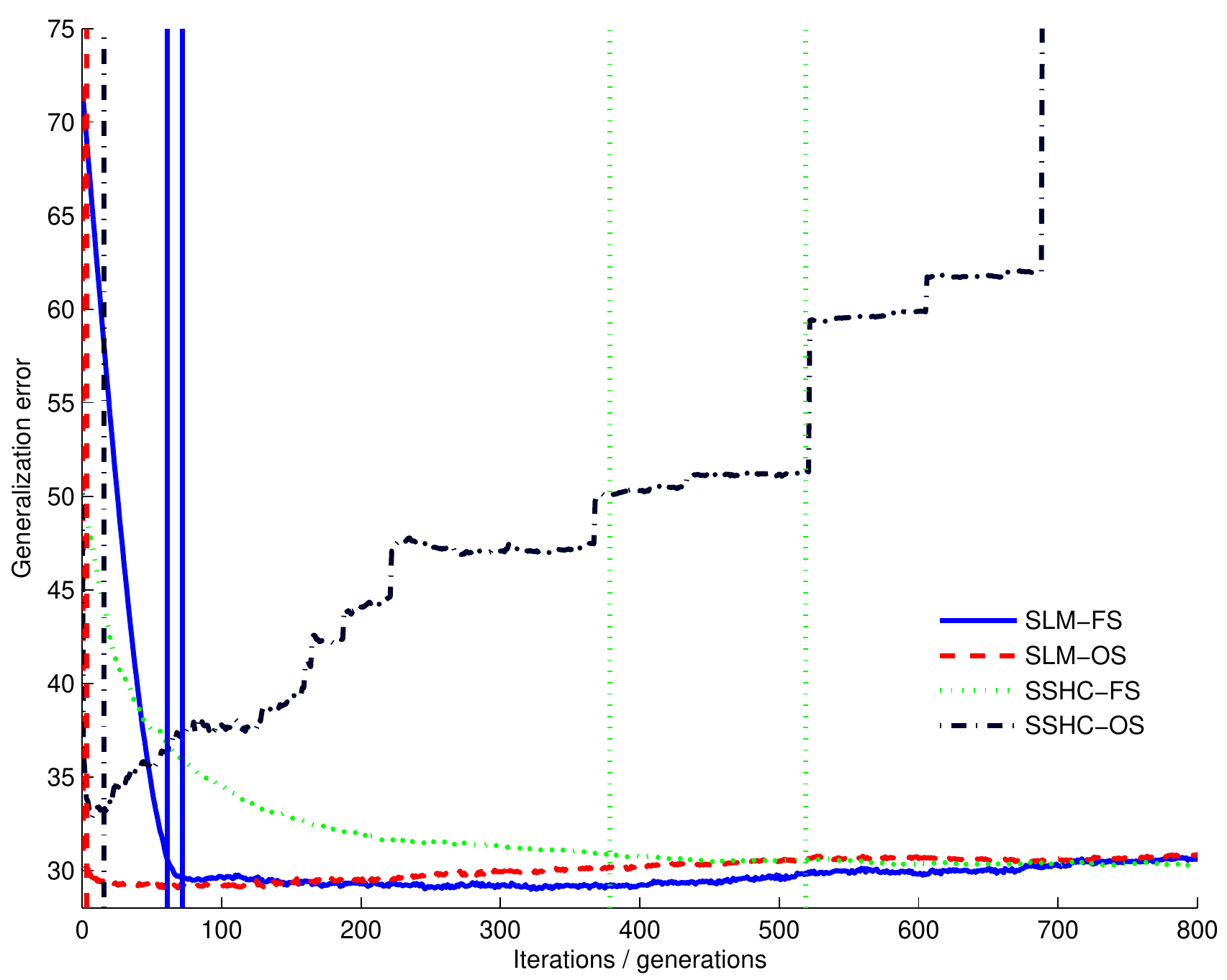}
		&
		\includegraphics[width=0.31\textwidth, keepaspectratio]{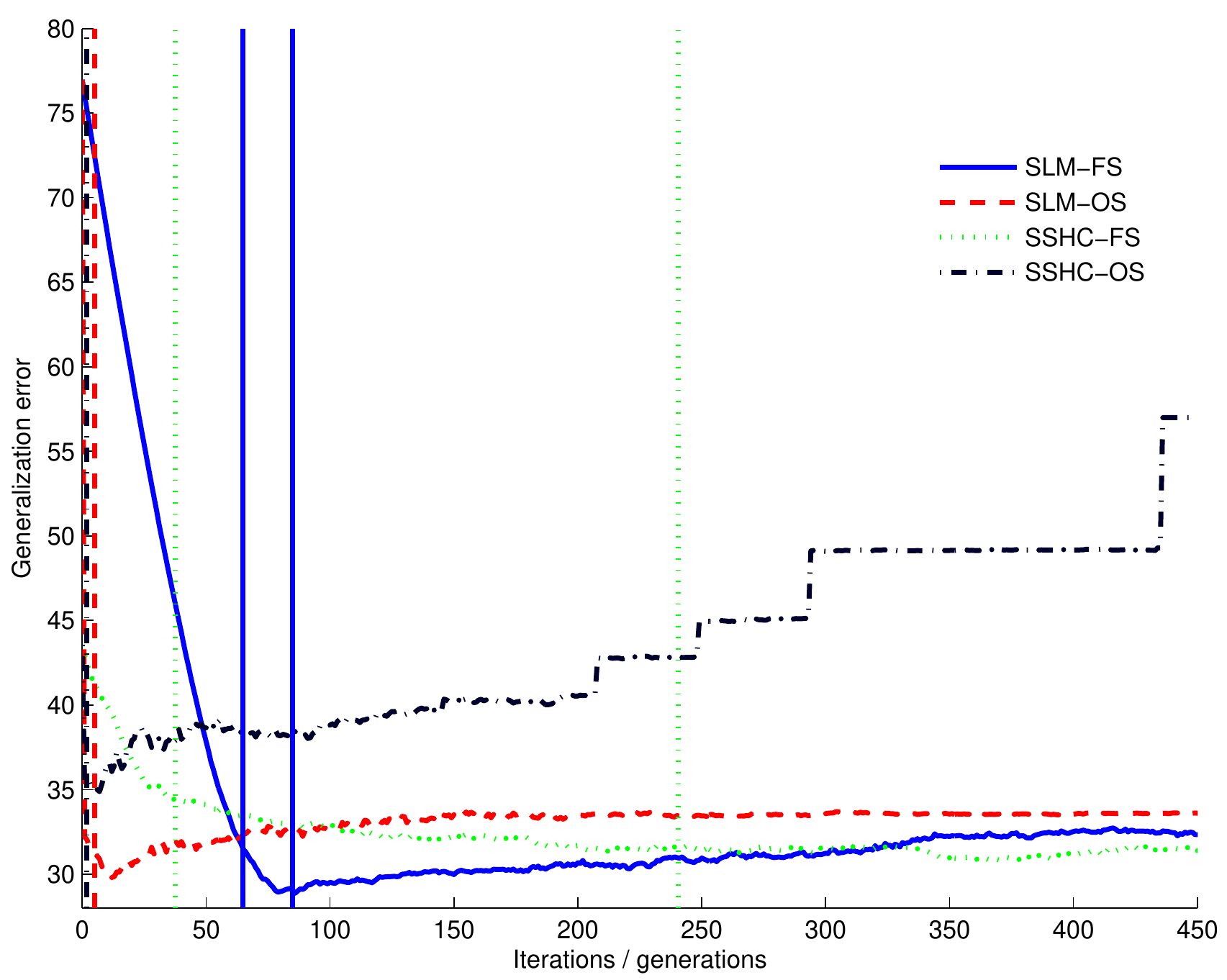}
		&
		\includegraphics[width=0.31\textwidth, keepaspectratio]{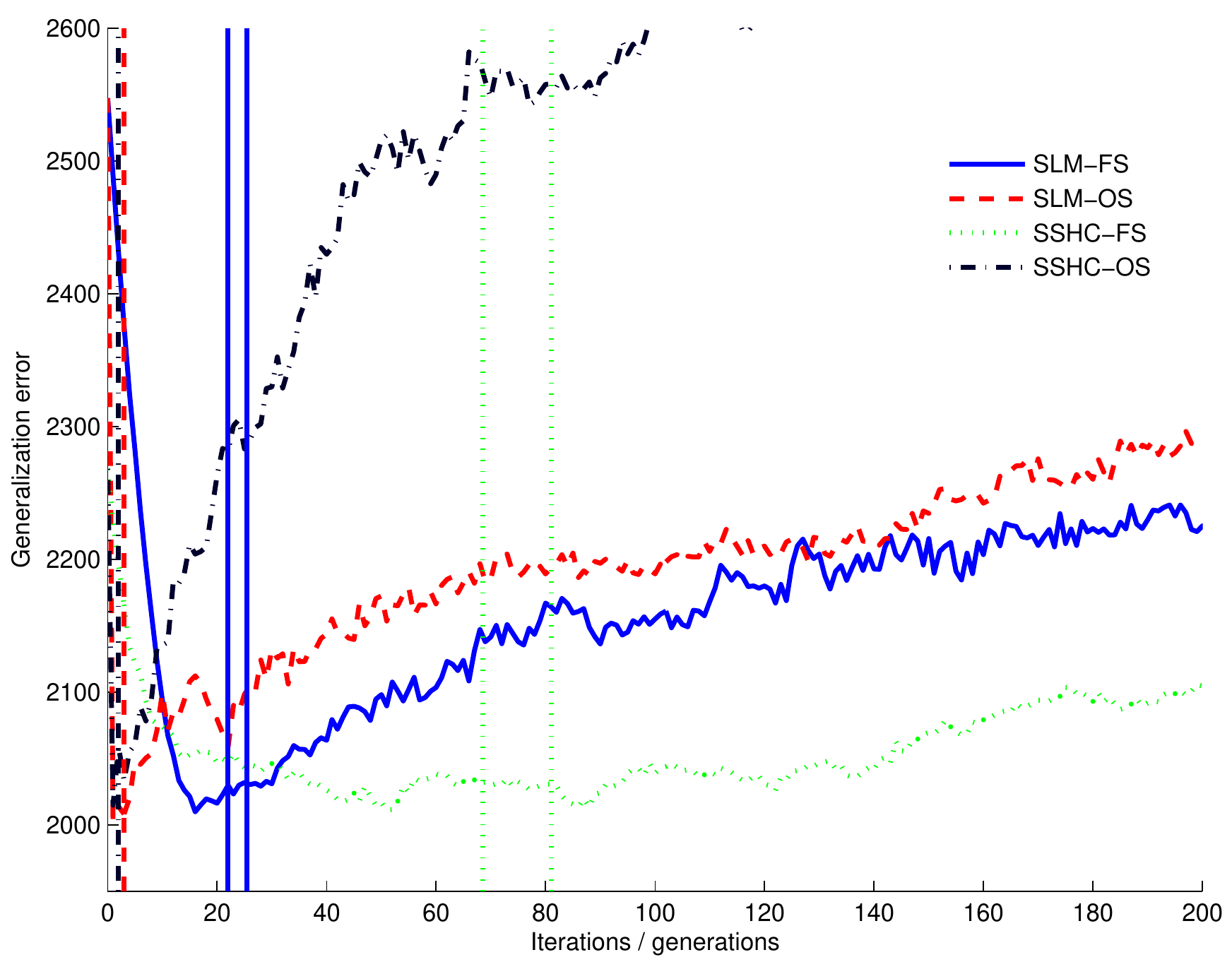}
		\\
		\includegraphics[width=0.31\textwidth, keepaspectratio]{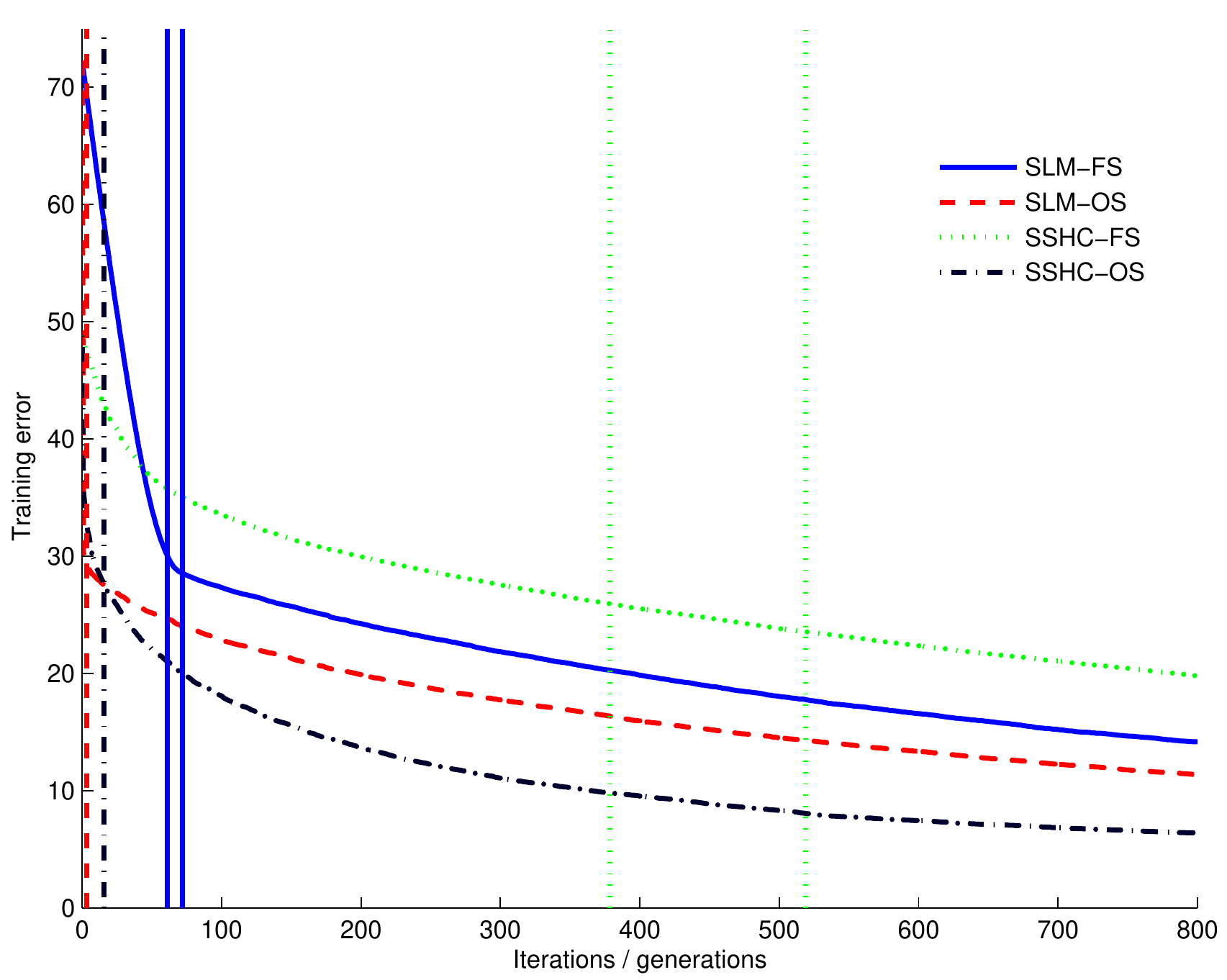}
		&
		\includegraphics[width=0.31\textwidth, keepaspectratio]{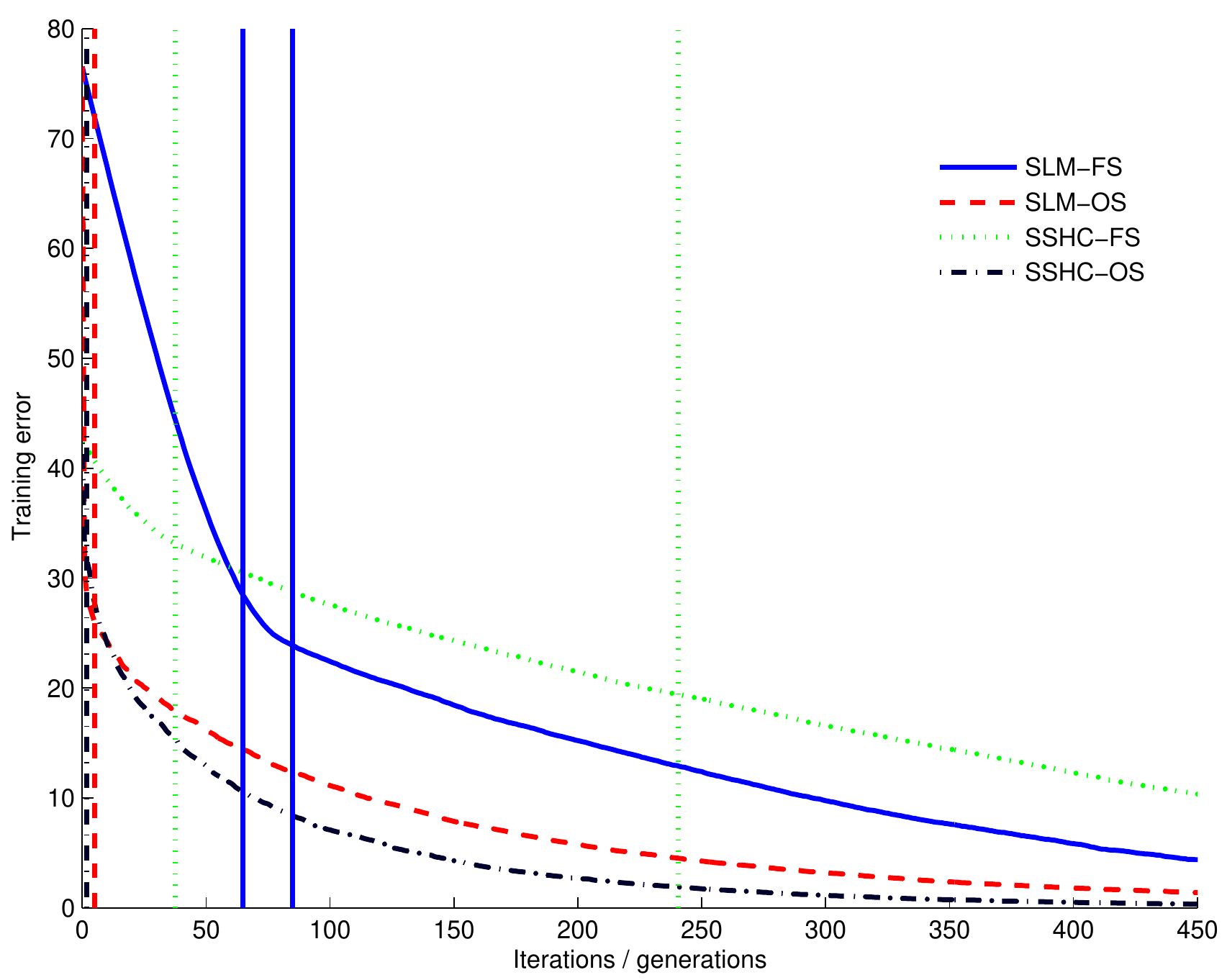}
		&
		\includegraphics[width=0.31\textwidth, keepaspectratio]{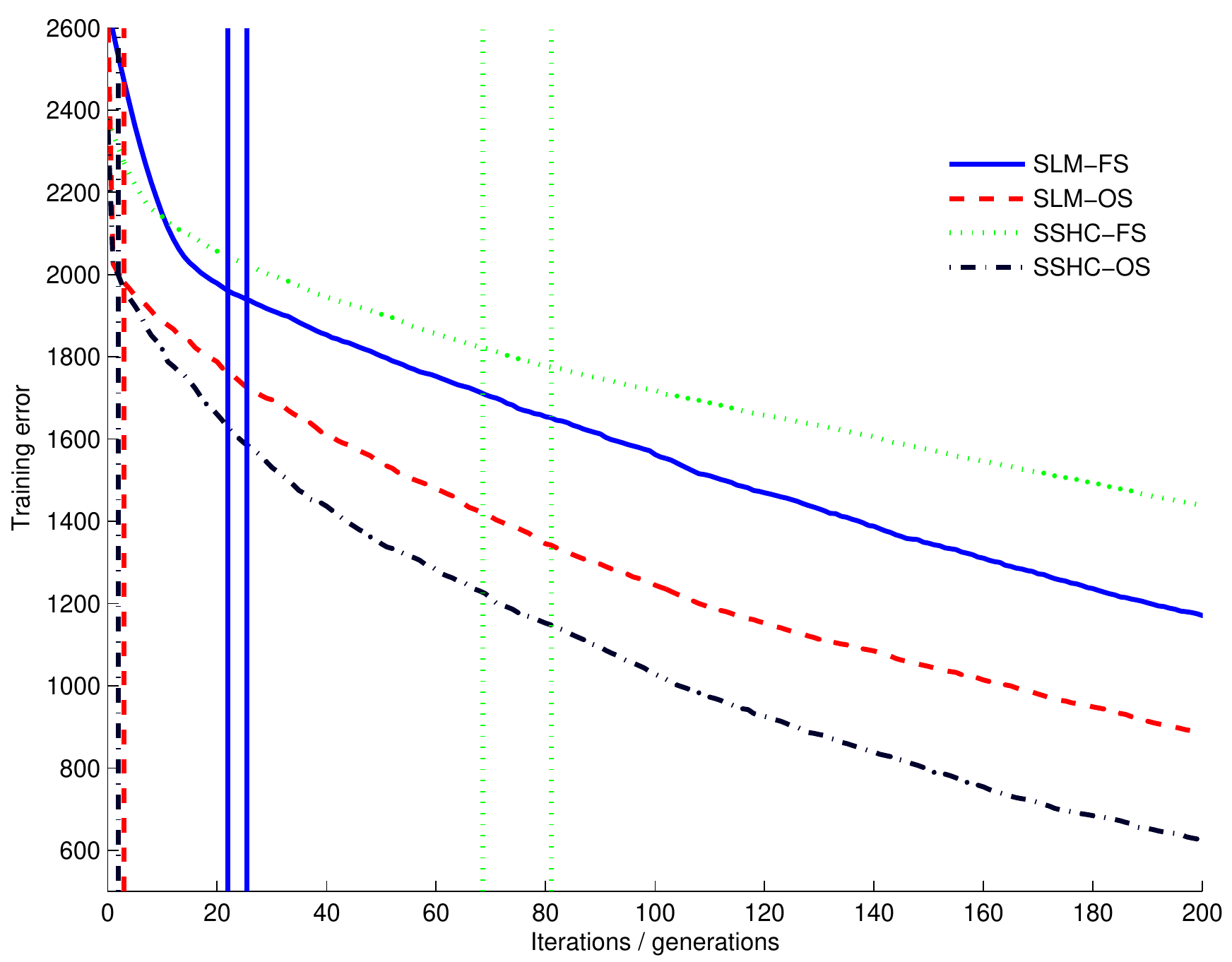}
	\end{tabular}
	\caption{Generalization (first row) and training (second row) errors evolution plots with the median stopping points represented as vertical lines for each corresponding variant. When both stopping criteria apply, the first stopping point represents the EDV stop, with the single exception occurring in the LD50 dataset where SLM-FS TIE stops earlier than SLM-FS EDV}
	\label{fig:comparison-plot}
\end{figure*}

In order to assess if the proposed stopping criteria do not stop too early, each variant is extended for more iterations/generations to determine if no further induction is indeed possible. Figure~\ref{fig:comparison-plot} presents the evolution plots with the generalization (first row) and training (second row) errors when no stopping is applied. The vertical lines represent the median stopping point for each variant. When both stopping criteria apply, the first stopping point represents the EDV stop, with the single exception occurring in the LD50 dataset where SLM-FS TIE stops earlier than SLM-FS EDV. The clear trend across all datasets and variants is that, after at least the best stopping point, the corresponding variant either starts to overfit or it simply stabilizes the generalization error. This hints that no further induction improvement is possible with the available data. Even if the generalization error stabilizes across the iterations/generations, there is no advantage in continuing the search as this would only increase model size and computational time. Notice also that after the stopping points, each corresponding variant is still able to effectively decrease the training error. This is particularly important to consider in the fixed step variants that use the TIE criterion, as this means that the step is not simply too high for the search to be effective. Therefore, the FS TIE variants are indeed detecting risky overfitting regions.
% perhaps extend this TIE part a little bit

% ##### 3) size #####

Table~\ref{tab:slm-size} presents the number of hidden neurons of the resulting Neural Networks for each SLM variant. Notice that since one hidden neuron is added at each successful iteration, the total number of hidden neurons is the number of iterations plus the initial random node. The resulting Neural Networks are relatively small, particularly for SLM-OS EDV with at most 8 hidden neurons on average. The results for the number of tree nodes of each SSHC variant are presented in table~\ref{tab:sshc-size}. The same reasoning applies here with SSHC-OS EDV producing small trees, in particular in the PPB and LD50 datasets. Notice that despite the considerable size of the trees produced by SSHC-FS TIE, this variant is able to surpass the other SSHC variants in terms of generalization in two out of the three datasets.
% Maybe add connection with Ensemble Learning and other paper here

\begin{table}
	\centering
	\caption{Median, average, and standard deviation (SD) outcomes for the number of hidden neurons of each SLM variant}
	\label{tab:slm-size}
	\resizebox{!}{18mm}{
	%\scalebox{0.75}{
	\begin{tabular}{c|c|c|c|c}
		\multirow{2}{*}{\bf{Dataset}} & \multirow{2}{*}{\bf{SLM variant}} & \multicolumn{3}{c}{\bf{Number of hidden neurons}} \\ \cline{3-5}
		& & \bf{Median} & \bf{Average} & \bf{SD} \\ \hline
		
		\multirow{3}{*}{\bf{Bio}}

		& \multirow{1}{*}{\bf{FS EDV}}
		& 62.0 & 61.6 & 2.1 \\ \cline{2-5}

		& \multirow{1}{*}{\bf{FS TIE}}
		& 73.0 & 73.3 & 1.7 \\ \cline{2-5}

		& \multirow{1}{*}{\bf{OS EDV}}
		& 4.0 & 4.2 & 1.1 \\ \hline

		\multirow{3}{*}{\bf{PPB}}

		& \multirow{1}{*}{\bf{FS EDV}}
		& 66.0 & 65.3 & 2.2 \\ \cline{2-5}

		& \multirow{1}{*}{\bf{FS TIE}}
		& 86.0 & 87.2 & 4.5 \\ \cline{2-5}

		& \multirow{1}{*}{\bf{OS EDV}}
		& 6.0 & 7.9 & 5.6 \\ \hline

		\multirow{3}{*}{\bf{LD50}}

		& \multirow{1}{*}{\bf{FS EDV}}
		& 26.5 & 28.6 & 7.6 \\ \cline{2-5}

		& \multirow{1}{*}{\bf{FS TIE}}
		& 23.0 & 23.2 & 2.4 \\ \cline{2-5}

		& \multirow{1}{*}{\bf{OS EDV}}
		& 4.0 & 4.9 & 2.6

	\end{tabular}}
\end{table}

\begin{table}
	\centering
	\caption{Median, average, and standard deviation (SD) outcomes for the number of tree nodes of each SSHC variant}
	\label{tab:sshc-size}
	\resizebox{!}{18mm}{
	%\scalebox{0.75}{
	\begin{tabular}{c|c|c|c|c}
		\multirow{2}{*}{\bf{Dataset}} & \multirow{2}{*}{\bf{SSHC variant}} & \multicolumn{3}{c}{\bf{Number of tree nodes}} \\ \cline{3-5}
		& & \bf{Median} & \bf{Average} & \bf{SD} \\ \hline
		
		\multirow{3}{*}{\bf{Bio}}

		& \multirow{1}{*}{\bf{FS EDV}}
		& 10170.0 & 9884.7 & 8257.3 \\ \cline{2-5}

		& \multirow{1}{*}{\bf{FS TIE}}
		& 14822.0 & 14613.1 & 3003.2 \\ \cline{2-5}

		& \multirow{1}{*}{\bf{OS EDV}}
		& 327.0 & 593.6 & 903.3 \\ \hline

		\multirow{3}{*}{\bf{PPB}}

		& \multirow{1}{*}{\bf{FS EDV}}
		& 1134.0 & 1979.2 & 2052.6 \\ \cline{2-5}

		& \multirow{1}{*}{\bf{FS TIE}}
		& 7815.0 & 7740.2 & 2357.8 \\ \cline{2-5}

		& \multirow{1}{*}{\bf{OS EDV}}
		& 39.0 & 56.5 & 48.6 \\ \hline

		\multirow{3}{*}{\bf{LD50}}

		& \multirow{1}{*}{\bf{FS EDV}}
		& 2131.0 & 2109.5 & 929.5 \\ \cline{2-5}

		& \multirow{1}{*}{\bf{FS TIE}}
		& 2509.0 & 2614.5 & 1080.7 \\ \cline{2-5}

		& \multirow{1}{*}{\bf{OS EDV}}
		& 57.0 & 90.9 & 86.4

	\end{tabular}}
\end{table}

% ##### 4) computational time #####

A final note on the computational time of the variants tested. Tables~\ref{tab:slm-time} and~\ref{tab:sshc-time} present the computational time in seconds of each run. The usage of the proposed semantic stopping criteria and the application of incremental evaluation at each mutation operator results in demonstrably efficient search procedures. The computational times presented are computed without any form of implicit or explicit parallelization. These results show that it is possible to evolve Neural Networks with competitive generalizations in less than 3 seconds on average. The GP variants can take at most around 10 seconds to compute. This efficiency can be particularly important in turning GP into a more widely used supervised learning method, given that GP is sometimes perceived as being relatively slow.

\begin{table}
	\centering
	\caption{Median, average, and standard deviation (SD) outcomes for the computational time (in seconds) of each SLM variant run}
	\label{tab:slm-time}
	\resizebox{!}{18mm}{
	%\scalebox{0.75}{
	\begin{tabular}{c|c|c|c|c}
		\multirow{2}{*}{\bf{Dataset}} & \multirow{2}{*}{\bf{SLM variant}} & \multicolumn{3}{c}{\bf{Computational time in seconds}} \\ \cline{3-5}
		& & \bf{Median} & \bf{Average} & \bf{SD} \\ \hline
		
		\multirow{3}{*}{\bf{Bio}}

		& \multirow{1}{*}{\bf{FS EDV}}
		& 1.6 & 1.6 & 0.1 \\ \cline{2-5}

		& \multirow{1}{*}{\bf{FS TIE}}
		& 1.9 & 1.9 & 0.1 \\ \cline{2-5}

		& \multirow{1}{*}{\bf{OS EDV}}
		& 0.3 & 0.4 & 0.4 \\ \hline

		\multirow{3}{*}{\bf{PPB}}

		& \multirow{1}{*}{\bf{FS EDV}}
		& 1.7 & 1.7 & 0.1 \\ \cline{2-5}

		& \multirow{1}{*}{\bf{FS TIE}}
		& 2.2 & 2.2 & 0.2 \\ \cline{2-5}

		& \multirow{1}{*}{\bf{OS EDV}}
		& 0.3 & 0.4 & 0.5 \\ \hline

		\multirow{3}{*}{\bf{LD50}}

		& \multirow{1}{*}{\bf{FS EDV}}
		& 2.8 & 2.9 & 0.7 \\ \cline{2-5}

		& \multirow{1}{*}{\bf{FS TIE}}
		& 2.6 & 2.7 & 0.7 \\ \cline{2-5}

		& \multirow{1}{*}{\bf{OS EDV}}
		& 0.2 & 0.3 & 0.4

	\end{tabular}}
\end{table}

\begin{table}
	\centering
	\caption{Median, average, and standard deviation (SD) outcomes for the computational time (in seconds) of each SSHC variant run}
	\label{tab:sshc-time}
	\resizebox{!}{18mm}{
	%\scalebox{0.75}{
	\begin{tabular}{c|c|c|c|c}
		\multirow{2}{*}{\bf{Dataset}} & \multirow{2}{*}{\bf{SSHC variant}} & \multicolumn{3}{c}{\bf{Computational time in seconds}} \\ \cline{3-5}
		& & \bf{Median} & \bf{Average} & \bf{SD} \\ \hline

		\multirow{3}{*}{\bf{Bio}}

		& \multirow{1}{*}{\bf{FS EDV}}
		& 7.8 & 7.1 & 5.9 \\ \cline{2-5}

		& \multirow{1}{*}{\bf{FS TIE}}
		& 10.0 & 10.3 & 2.5 \\ \cline{2-5}

		& \multirow{1}{*}{\bf{OS EDV}}
		& 0.9 & 1.3 & 1.7 \\ \hline

		\multirow{3}{*}{\bf{PPB}}

		& \multirow{1}{*}{\bf{FS EDV}}
		& 0.3 & 0.6 & 0.6 \\ \cline{2-5}

		& \multirow{1}{*}{\bf{FS TIE}}
		& 2.1 & 2.1 & 0.8 \\ \cline{2-5}

		& \multirow{1}{*}{\bf{OS EDV}}
		& 0.1 & 0.1 & 0.2 \\ \hline

		\multirow{3}{*}{\bf{LD50}}

		& \multirow{1}{*}{\bf{FS EDV}}
		& 1.1 & 1.1 & 0.5 \\ \cline{2-5}

		& \multirow{1}{*}{\bf{FS TIE}}
		& 1.2 & 1.3 & 0.5 \\ \cline{2-5}

		& \multirow{1}{*}{\bf{OS EDV}}
		& 0.2 & 0.3 & 0.4

	\end{tabular}}
\end{table}

% ####################################################################################################
%
% Section: Conclusions
%
% ####################################################################################################
\section{Conclusions}

This paper showed that it is feasible to use information gathered from the semantic neighborhood to determine search stopping points that result in competitive generalizations. The semantic stopping criteria proposed are directly applicable to GSGP and to the SLM algorithm. Besides achieving competitive generalizations, the proposed approach also yields computationally efficient algorithms as it allows the evolution of neural networks in less than 3 seconds on average, and of GP trees in at most 10 seconds. This paper also explored for the first time the computation of optimal learning steps under the SLM algorithm. This results in the successful evolution of neural networks in just a few iterations. The resulting networks also have a small number of hidden neurons. In GSGP, the usage of the proposed semantic stopping criteria in conjunction with the computation of optimal mutation steps also results in small trees.

As future work, an extended experimental study will help to assess the general applicability of the proposed stopping criteria. This extended experimental assessment should include other regression datasets, as well as classification datasets, and it should also include comparisons with other well-established non-evolutionary supervised learning methods.

% ? maybe mention the following: An hybrid combination between using fixed and optimal steps is also in order.
%hybrid between fixed and optimal step given ...

% ? maybe mention validation set approach

%-from-thesis-start
%Experimentally assessing the SLM for the case of several hidden layers is one of the first future steps. It remains to be seen if the added hidden layers can provide a superior generalization or faster convergence, even thought in terms of representation, the single hidden layer case is sufficient under the universal approximation results. It is also important to study if the effectiveness of the semantic stopping criteria remains unaltered in the case of several hidden layers. The geometric semantic mutation defined for feedforward NNs can also be extended for the case of recurrent NNs. This extension is possible under mild restrictions. The corresponding experimental assessment of the geometric semantic mutation for recurrent NNs is also an important future step.
%-from-thesis-end

%\end{document}  % This is where a 'short' article might terminate

\begin{acks}
This work was partially funded by project PERSEIDS (PTDC/EMS-SIS/0642/2014) and BioISI RD unit, UID/MULTI/04046/2013, funded by FCT/MCTES/PIDDAC, Portugal. Funding for this work was also provided by CONACYT project FC-2015-2/944 "Aprendizaje evolutivo a gran escala".
\end{acks}

%\begin{acks}
  %The authors would like to thank Dr. Yuhua Li for providing the
  %matlab code of  the \textit{BEPS} method. 
%
  %The authors would also like to thank the anonymous referees for
  %their valuable comments and helpful suggestions. The work is
  %supported by the \grantsponsor{GS501100001809}{National Natural
    %Science Foundation of
    %China}{http://dx.doi.org/10.13039/501100001809} under Grant
  %No.:~\grantnum{GS501100001809}{61273304}
  %and~\grantnum[http://www.nnsf.cn/youngscientsts]{GS501100001809}{Young
    %Scientsts' Support Program}.
%
%\end{acks}

\bibliographystyle{ACM-Reference-Format}
\bibliography{sigproc} 

\end{document}